\documentclass[10pt,twocolumn,letterpaper]{article}

\usepackage{iccv}
\usepackage{times}
\usepackage{epsfig}
\usepackage{graphicx}
\usepackage{amsmath}
\usepackage{amssymb}
\usepackage{bbm}
\usepackage{soul}
\usepackage[accsupp]{axessibility}  % Improves PDF readability for those with disabilities.

\usepackage{float}
\usepackage{pifont}% http://ctan.org/pkg/pifont
\usepackage[breaklinks=true,bookmarks=false]{hyperref}

\iccvfinalcopy % *** Uncomment this line for the final submission

\def\iccvPaperID{****} % *** Enter the ICCV Paper ID here

\ificcvfinal\fi

\begin{document}

%%%%%%%%% TITLE
\title{BiMaL: Bijective Maximum Likelihood Approach to \\Domain Adaptation in Semantic Scene Segmentation}

\author{Thanh-Dat Truong$^{1}$, Chi Nhan Duong$^{2}$, Ngan Le$^{1}$, Son Lam Phung$^{4}$, Chase Rainwater$^{3}$, Khoa Luu$^{1}$\\
$^{1}$CVIU Lab, University of Arkansas \quad 
$^{2}$Concordia University \\
$^{3}$Dep. of Industrial Engineering, University of Arkansas \quad
$^{4}$University of Wollongong\\
% Institution1 address\\
{\tt\small \{tt032, thile, cer, khoaluu\}@uark.edu, dcnhan@ieee.org, phung@uow.edu.au}
% For a paper whose authors are all at the same institution,
% omit the following lines up until the closing ``}''.
% Additional authors and addresses can be added with ``\and'',
% just like the second author.
% To save space, use either the email address or home page, not both
% \and
% Second Author\\
% Institution2\\
% First line of institution2 address\\
% {\tt\small secondauthor@i2.org}
}

\maketitle
% Remove page # from the first page of camera-ready.
\ificcvfinal\thispagestyle{empty}\fi

%%%%%%%%% ABSTRACT
\begin{abstract}
Semantic segmentation aims to predict pixel-level labels. It has become a popular task in various computer vision applications. While fully supervised segmentation methods have achieved high accuracy on large-scale vision datasets, they are unable to generalize on a new test environment or a new domain well. In this work, we first introduce a new Unaligned Domain Score to measure the efficiency of a learned model on a new target domain in unsupervised manner. Then, we present the new Bijective Maximum Likelihood\footnote{\url{https://github.com/uark-cviu/BiMaL}} (BiMaL) loss that is a generalized form of the Adversarial Entropy Minimization without any assumption about pixel independence. We have evaluated the proposed BiMaL on two domains. The proposed BiMaL approach consistently outperforms the SOTA methods on empirical experiments on ``SYNTHIA to Cityscapes'', ``GTA5 to Cityscapes'', and ``SYNTHIA to Vistas''.

\end{abstract}

%%%%%%%%% BODY TEXT
\section{Introduction}

Semantic segmentation is one of the most popular  computer vision topics, which aims to to assign each pixel in an image to a 
% corresponding 
predefined class.
It has various practical applications, especially in autonomous driving where a segmentation model is needed to recognize roads, sidewalks, pedestrians or vehicles in a large variety of urban conditions. A typical supervised segmentation model is usually trained on datasets with labels. However, annotating images for the semantic segmentation task is costly and time-consuming. 
% Instead, a common feasible
Alternatively, a powerful and cost-effective way to acquire a large-scale training set is to use a simulation, e.g. game engines, to create a synthetic dataset \cite{Richter_2016_ECCV, Ros_2016_CVPR}. %[SYNTHIA, GTA].  
However, fully supervised models \cite{chen2018deeplab, le2018segmentation} trained on the synthetic datasets are often unable to perform well on real images due to the pixel appearance gap between synthetic and real images.

% \begin{figure}[t]
%     \centering
%     \includegraphics[width=0.45\textwidth]{figures/figure1.png}
%     \caption{Fig 1.}
%     \label{fig:same_entropy_1}
% \end{figure}

\begin{figure*}[!t]
    \centering
    \includegraphics[width=0.95\textwidth]{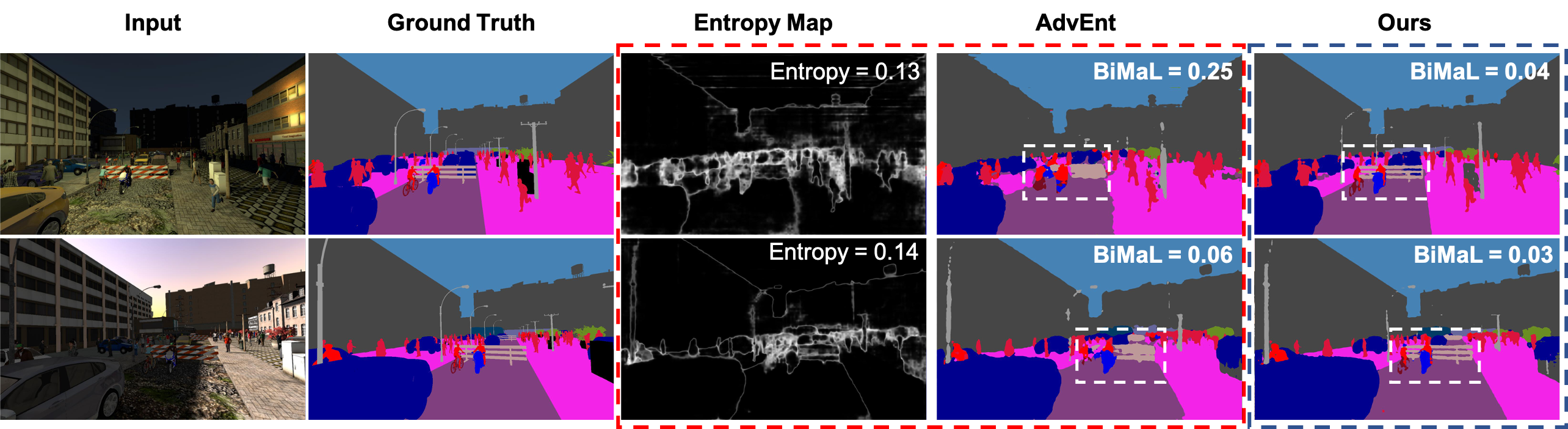}
    \caption{\textbf{Two images have the same entropy but one has a poor prediction (a top image) and one has an better prediction (a bottom image).} Columns 1 and 2 are an input image and a ground truth. Columns 3 and 4 are an entropy map and a prediction of AdvEnt \cite{vu2019advent}. Column 5 is the results of our proposed method. The two predictions produced by AdvEnt have similar entropy scores ($0.13$ and $0.14$). Meanwhile, the BiMaL value of the bottom prediction ($0.06$) is smaller than the top prediction ($0.14$). Our results in the last column, which have better BiMaL values than AdvEnt, can well model the structure of an image. In particular, our results have sharper results of a barrier and a rider (white dash box), and a clear boundary between road and sidewalk.}
    % The left column includes an RGB input and corresponding ground truth (GT) image. The right columns includes an entropy map and predicted segmentation. These two images have very close entropy scores ($0.13$ and $0.14$). Meanwhile, our proposed BiMaL value gives a small value of correct prediction ($0.06$) and a high value of an incorrect prediction ($0.25$). The lower BiMaL value is the better prediction. 
    %\hl{incorrect results were conducted by [?]. what is resutls by BiMaL loss}} % \hl{Can we have our results here to compare?}}
    \label{fig:same_entropy_1}
    \vspace{-4mm}
\end{figure*}

Unsupervised Domain Adaptation (UDA) aims to train a machine learning model on an annotated dataset, i.e. the source, and guarantee its high performance on a new unlabeled dataset, i.e. the target.
The UDA approaches have been applied to various computer vision tasks such as Semantic Segmentation \cite{chen2018deeplab, le2018segmentation, li2020content, vu2019advent, vu2019dada, zhang2019category}, Face Recognition \cite{duong2019shrinkteanet, Luu_BTAS2009, Luu_FG2011, Luu_ROBUST2008, Luu_IJCB2011}. 
The recent UDA methods aim to reduce the cross-domain discrepancy, along with the supervised training on the source domain \cite{chen2018road, ganin2015unsupervised, long2015learning,pan2020unsupervised, tzeng2017adversarial, vu2019advent}.
In particular, these methods aim to minimize the distribution discrepancy of the 
deep representations extracted from the source and the target domains. This process can be performed at single or multiple levels of deep features using maximum mean discrepancies \cite{ganin2015unsupervised, long2015learning, tzeng2017adversarial}, or adversarial training \cite{chen2018road, chen2017no, hoffman18a, hoffman2016fcns, hong2018CVPR, tsai2018learning}. The approaches in this group have shown their potential in aligning the predicted outputs of images from the two domains. However, 
%and constraining on the overall structure of the predicted segmentation maps, 
the binary cross-entropy 
% \hl{It doesnt seem that. It seems to me "Categorical CE" is used for multiple classes $\rightarrow$ \textbf{this point mentions about the discriminator in GAN to classify the output from src or target domain}}
label predicted by the learned discriminator is usually a weak indication of structural learning for the segmentation task. Another approach named self-training utilizes the pseudo-labels or generative networks conditioned on target images \cite{murez2018CVPR, zhu2017unpaired}. Semi-supervised learning is an approach related to UDA where the training set consists of both labeled and unlabeled samples. Thus, it has motivated several UDA approaches such as Class-balanced self-training (CBST) \cite{zou2018unsupervised},  %maximum likelihood \cite{7472651}, 
and entropy minimization \cite{chen2019domain, grandvalet2005semi, pan2020unsupervised, springenberg2015unsupervised, vu2019advent}.
Although metrics such as entropy can be efficiently computed and adopted for training, they tend to rely on easy predictions, i.e. high confident scores, as references for the label transfer from source to target domains. This issue is alleviated in a later approach \cite{chen2019domain} by preventing learned models from over-focusing on high confident areas. However, this type of metrics is formulated in pixel-wised manner, and, therefore, neglects the structural information presented in the image (see Figure \ref{fig:same_entropy_1}). 

\noindent
\textbf{Contributions of this Work. }
This work presents a new unsupervised domain adaptation approach to tackle the semantic segmentation problem. Table \ref{tab:summary} summarizes the difference between our proposed approach and the prior ones. Our contributions can be summarized as follows.

Firstly, a new Unaligned Domain Score (UDS) is introduced to measure the efficiency of the learned model on a target domain in an unsupervised manner.
Secondly, the presented UDS is further extended as a new loss function, named Bijective Maximum Likelihood (BiMaL) loss, that can be used with an unsupervised deep neural network to generalize on target domains. Indeed, we further demonstrate BiMaL loss is a generalized form of the Adversarial Entropy Minimization (AdvEnt) \cite{vu2019advent} without pixel independence assumption. 
Far apart from AdvEnt that assumes pixel independence, BiMaL loss is formed using a Maximum-likelihood formulation to model the global structure of a segmentation input and a Bijective function to map that segmentation structure to a deep latent space.
Finally, the proposed BiMaL method is evaluated on three popular large-scale semantic segmentation benchmarks, including GTA5 \cite{Richter_2016_ECCV} $to$ CityScapes \cite{cordts2016cityscapes}, SYNTHIA \cite{Ros_2016_CVPR} $to$ Cityscapes, and SYNTHIA $to$ Vista \cite{MVD2017}. The experimental results demonstrate our proposed BiMaL approach consistently outperforms the State-of-the-Art (SOTA) methods \cite{chen2018road, pan2020unsupervised, tsai2018learning, vu2019advent, vu2019dada} in all these benchmark databases.
% \hl{
To the best of our knowledge, this is one of the first works that introduces a novel bijective maximum likelihood approach with flow-based metric to unsupervised domain adaptation in semantic segmentation.
% }

% %novel bijective maximum likelihood approach, derived from flow-based generative model, 
% to unsupervised domain adaptation in semantic segmentation.
% }

% I think the author can emphasis more on the novelty and groundbreaking of the paper. Because recent papers can be divided to two main directions: domain adversarial learning and self-training. This paper introduces a new framework: flow-based model for UDA. However, I think the writing of the abstract and introduction are too plain.

\section{Related Work}

Unsupervised Domain Adaptation has recently become one of the most active research topics. The common UDA approaches are domain discrepancy minimization \cite{ganin2015unsupervised, long2015learning, tzeng2017adversarial}, adversarial learning \cite{chen2018road,chen2017no, hoffman18a, hoffman2016fcns, hong2018CVPR, tsai2018learning}, entropy minimization \cite{murez2018CVPR, pan2020unsupervised, vu2019advent, zhu2017unpaired}, self-training \cite{zou2018unsupervised}. In the scope of this work, UDA is focused on semantic segmentation.
% Since we target on semantic segmentation in this work, we just give a review of UDA approaches that aims to our task.

Adversarial training is the most common approach employed to UDA for semantic segmentation. Similar to generative adversarial networks (GANs), the adversarial training paradigm aims at training a discriminator to predict the domain of inputs while the segmentation network tries to fool the discriminator. 
This adversarial step is trained simultaneously with the supervised segmentation task on the source domain.
% This adversarial step also goes along with the supervised segmentation task on the source domains. 
Hoffman \etal \cite{hoffman2016fcns} first introduced GAN-based UDA approach to semantic segmentation. Later, Chen \etal \cite{chen2017no} presented global and class-wise adaptation learned by adversarial learning on pseudo labels. Considering the difference in spatial distribution, \cite{chen2018road} proposed a spatial-aware adaptation method to align two domains along with a target guided distillation loss. Hong \etal \cite{hong2018CVPR} learned a conditional generator to transform the feature maps of source domain to be similar to target domain. 
Tasi \etal \cite{tsai2018learning} used adversarial learning to learn a consistency of scene layout and local context between source and target domains. There are some prior methods that utilize the generative networks to synthesize target images conditioned on source images \cite{zhu2017unpaired, murez2018CVPR}. Hoffman \etal \cite{hoffman18a} presented Cycle-Consistent Adversarial Domain Adaptation that aligns at both pixel-level and feature-level representations. Zhu~\etal \cite{zhu2018ECCV} introduced a conservative loss in an adversarial framework that penalizes the easy and hard source examples. We \etal \cite{wu2018dcan} proposed a DCAN framework that uses the channel-wise feature alignment in the segmentation networks.
Sakaridis \etal \cite{SDHV18} proposed an UDA framework on scene understanding that gradually adapts a segmentation model 
from non-foggy to heavy-foggy images.

\begin{table*}[t]
% \small
\centering
\caption{ \textbf{Comparison in the properties between our proposed approach and other methods}. Convolutional Neural Network (CNN), Generative Adversarial Net (GAN), Bijective Network (BiN), Entropy Minimization (EntMin), Curriculum Training (CT), Image-wise Weighting (IW), Segmentation Map (Seg), Depth Map (Depth); $\ell_{CE}$: Cross-entropy Loss, $\ell_{adv}$: Adversarial Loss, $\ell_{Huber}$: Huber Loss.}
% ; $-$ indicates ``No'' or ``Not applicable''. 
%\hl{it seems to me that there lacks of methods in unsupervised segmentation }}
\small
\begin{tabular}{|c|c|c|c|c|c|}
\hline
%\textbf{Methods} & \textbf{Architecture}              & \textbf{Privileged   Information} & \textbf{Loss Function}                       & \textbf{Global   Structural Guarantee} \\ \hline
\textbf{Methods} & \textbf{Architecture}              & \textbf{Source Label} & \textbf{Learning Mechanism} & \textbf{Loss Function}                       & \textbf{Structural Learning} \\ \hline
AdaptSeg \cite{tsai2018learning}        & CNN + GAN                          & Seg  & Domain Adaptation & $\ell_{adv}$                                 & Weak (binary label)                                     \\ \hline
AdaptPatch \cite{tsai2019domain}     & CNN + GAN                          & Seg & Domain Adaptation & $\ell_{adv}$                                  & Weak (binary label)                                     \\ \hline
CBST \cite{zou2018unsupervised}             & CNN                                & Seg &           Self-Training                    &       $\ell_{CE}$                          & Not Applicable                                    \\ \hline
ADVENT \cite{vu2019advent}          & CNN + GAN                          & Seg   & Domain Adaptation & EntMin                         & Weak (binary label)                                     \\ \hline
MaxSquare \cite{chen2019domain}         & CNN + GAN                          & Seg  & Domain Adaptation & Squares loss + IW & Weak (binary label)                                     \\ \hline
IntraDA \cite{pan2020unsupervised}         & CNN + GAN                          & Seg  & Curriculum Learning & EntMin & Weak (binary label)                                     \\ \hline \hline
SPIGAN \cite{lee2018spigan}         & CNN + GAN                          & Seg + Depth & Domain Adaptation & $\ell_{adv}$ + $\ell_{1}$                          & Weak (binary label)                                    \\ \hline
DADA \cite{vu2019dada}            & CNN + GAN                          & Seg + Depth  & Domain Adaptation & $\ell_{adv}$ + $\ell_{Huber}$                          & Depth-aware Label                                   \\ 
\hline \hline
\textbf{BiMaL}  & \textbf{CNN + BiN} & \textbf{Seg}  & \textbf{Domain Adaptation} & \textbf{Maximum   Likelihood}      & \begin{tabular}{@{}c@{}} \textbf{Segmentation Density} \\ \textbf{(Unsupervised)}\end{tabular}                           \\ \hline
\end{tabular}
\label{tab:summary}
\vspace{-4mm}
\end{table*}

% In order to
To enhance the performance of domain adaptation, several methods explore the use of privileged information available on source data \cite{chen2014recognizing, li2014exploiting, sarafianos2017adaptive}. Vapnik \etal \cite{vapnik2009new} first introduced the idea of privileged information, i.e. additional information only available at the training process. Later, many methods \cite{hoffman2016learning, lopez2015unifying, mordan2018revisiting, Sharmanska_2013_ICCV} take advantage of privileged information for various tasks. In semantic segmentation, SPIGAN \cite{lee2018spigan} proposed an UDA approach that utilizes the depth information during the training phase. Following SPIGAN,  Vu \etal\cite{vu2019dada} presented an adversarial approach that utilizes the depth-aware of source and target images.

% Curriculum Domain Adaptation. Our work is also related to curriculum domain adaptation [22, 31, 7] which
% deals with easy samples first. For curriculum domain adaptation on scene understanding, [22] proposes to adapt
% a semantic segmentation model from non-foggy images to
% synthetic light foggy images, and then to real heavy foggy
% images. To generalize this concept, [7] decomposes the domain discrepancy into multiple smaller discrepancies by introducing unlabeled intermediate domains. However, these
% techniques require additional information to decompose domains. To cope with this limitation, [31] focuses on learning the global and local label distributions of images as the
% first task to regularize the model predictions in the target
% domain. In contrast, we propose a simpler and data-driven
% approach to learn the easy target samples based on an entropy ranking system.

Entropy minimization has been used for semi-supervised learning \cite{grandvalet2005semi, springenberg2015unsupervised}. Vu et al. \cite{vu2019advent} first introduced the entropy minimization approach for domain adaptation in semantic segmentation. The minimization process is solved by adversarial learning. Later, \cite{pan2020unsupervised} introduced an intra-domain adaptation approach based on the entropy level of predictions. The learning process involves two phases. The first phase performs adaptation from the source domain to the target domain, whereas the second phase aligns the hard split and easy split within the target domain.
Another recent UDA approach is self-training, where 
% the main idea is to use 
the predictions of the trained model are used as pseudo-labels for the unlabeled data to train the new model. Self-training has been widely used in classification \cite{NEURIPS2019_bf25356f}
%. This approach also benefits for the 
 and segmentation tasks \cite{zou2018unsupervised}. 

%\textcolor{red}{What are the limitations of these prior approaches? Then link to the next section.}

% Another approach for UDA is self-training. The idea is
% to use the prediction from an ensembled model or a previous state of model as pseudo-labels for the unlabeled data
% to train the current model. Many semi-supervised methods
% [20, 39] use self-training. In [51], self-training is employed
% for UDA on semantic segmentation which is further extended with class balancing and spatial prior. Self-training
% has an interesting connection to the proposed entropy minimization approach as we discuss in Section 3.1.

% Contrastive Adaptation Network (CAN Network):   %\url{https://arxiv.org/pdf/1901.00976.pdf}

%  DADA: Depth-Aware Domain Adaptation % \url{https://arxiv.org/pdf/1904.01886.pdf}

% ADVENT: Adversarial Entropy Minimization % \url{https://arxiv.org/pdf/1811.12833.pdf}

% Unsupervised Intra-Domain Adaptation  % \url{https://arxiv.org/pdf/2004.07703.pdf}

% \section{The Proposed Method}

% Using Conditional Random Field to model the output segmentation

% \begin{equation}
% % \footnotesize
% \begin{split}
%     \mathcal{L} &= \sum_i\psi(x_i) + \sum_{i,j}\psi(x_i, x_j) \\
%                 &= \sum_i\log(p(x_i)) \\
%                 &+ \sum_{i,j}\left(-w^{i,j}_1\frac{(p(x_i) - p(x_j))^2}{2\sigma_1^2} -w^{i,j}_2\frac{(u_i - u_j)^2}{2\sigma_2^2}\right) \\
%                 &= \left|\frac{\partial f(\mathbf{x})}{\partial \mathbf{x}}\right|^{-1}\log(p(z_i)) \\ 
%                 &+ \sum_{i,j}\left(-w^{i,j}_1\frac{(p(z_i) - p(z_j))^2}{2\sigma_1^2} -w^{i,j}_2\frac{(u_i - u_j)^2}{2\sigma_2^2}\right)\\
% \end{split}
% \end{equation}

%\section{Domain Disordered Scores (DDS)}
\section{Unaligned Domain Scores (UDS)}

% In this section, 
% This section presents a new Unaligned Domain Score to evaluate the robustness of a segmentation model. %predicted segmentation model. 
% We firstly overview the unsupervised domain adaptation with the direct entropy minimization. Although the standard entropy minimization loss appropriates for unsupervised learning, it is still unable to model the global structure of a semantic segmentation due to its assumption on pixel independence.
% Then, a new Unaligned Domain Score will be defined. It is used to measure the efficiency of the trained segmentation model on a target dataset. This UDS score is further extended to a new unsupervised loss function presented in Section \ref{sec:maximum_likelihood}.

% \subsection{Overview of Unsupervised Domain Adaptation with Direct Entropy Minimization}

\begin{figure*}[!t]
    \centering
    \includegraphics[width=0.8\textwidth]{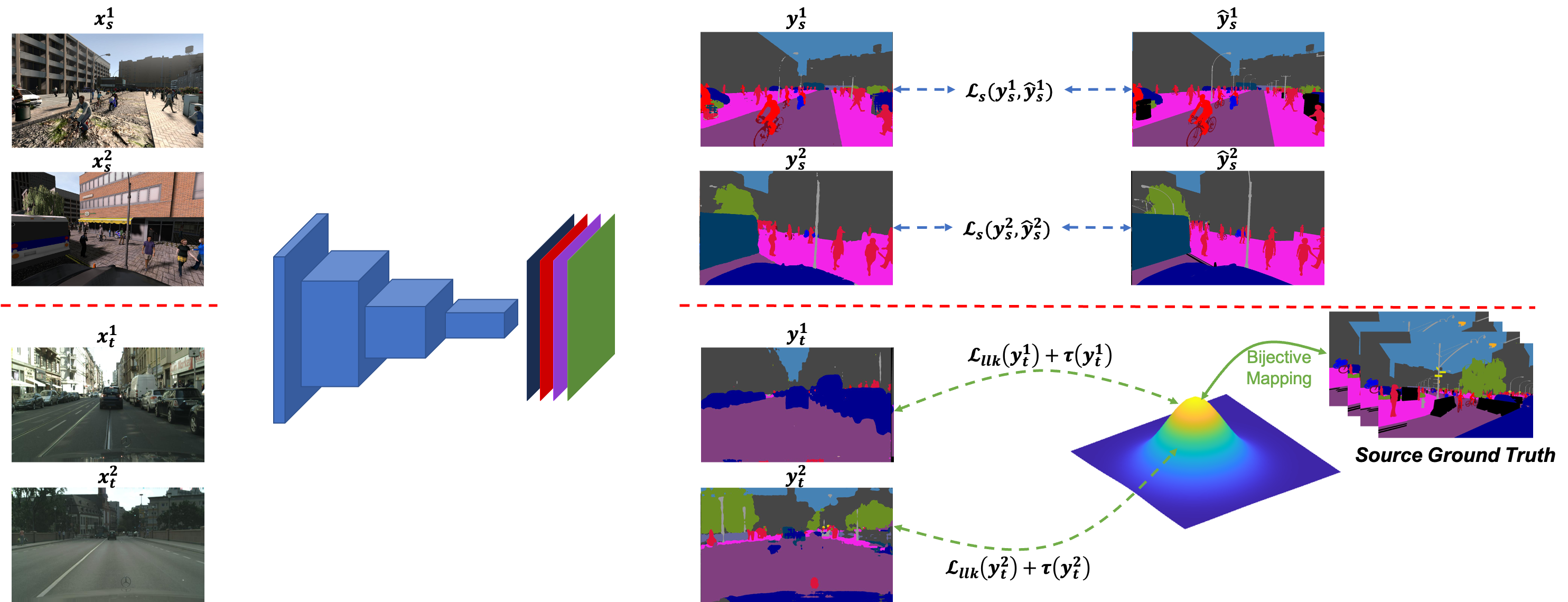}
    \caption{\textbf{The Proposed Framework.} The RGB image input is firstly forwarded to a deep semantic segmentation network to produce a segmentation map. The supervised loss is employed on the source training samples. Meanwhile, the predicted segmentation on target training samples will be mapped to the latent space to compute the Bijective Maximum Likelihood loss. The bijective mapping network is trained on the ground-truth images of the source domain.}
    \label{fig:proposed_framework}
\end{figure*}

Let $\mathbf{x}_s \in \mathcal{X}_s \subset \mathbb{R}^{H \times W \times 3}$ be an input image of the source domain ($H$ and $W$ are the height and width of an image),  $\mathbf{x}_t \in \mathcal{X}_t \subset \mathbb{R}^{H \times W \times 3}$ be an input image of the target domain, $G: \mathcal{X} \to \mathcal {Y}$ where $\mathcal{X} = \mathcal{X}_s \cup \mathcal{X}_t$ be a semantic segmentation function 
%the deep network segmentation 
that maps an input image to its corresponding segmentation map $\mathbf{y} \subset \mathbb{R}^{H \times W \times C}$, i.e. $\mathbf{y} = G(\mathbf{x}, \theta)$ ($C$ is the number of semantic classes). 
% Let $p(\mathbf{y})$ be a probability density function of predicted segmentation with the corresponding input $\mathbf{x}$ (i.e. $\mathbf{y} = g(\mathbf{x})$). 
In general, given $N$ labeled training samples from a source domain 
$\mathcal{D}_s = \{\mathbf{x}_s^i, \hat{\mathbf{y}}_s^i\}_1^N$ 
%$\mathcal{D}_s = \mathcal{X}_s \times \mathcal{Y}_s$ 
and $M$ unlabeled samples from a target domain $\mathcal{D}_t = \{\mathbf{x}_t^i\}_1^M$, the 
unsupervised domain adaptation for semantic segmentation can be formulated as: 
% in Eqn. \eqref{eqn:objective}, %follows,
%aims to optimize the network $g$ respecting to the condition as shown in Eqn. \eqref{eqn:objective}.
% \begin{equation}
%     g^{*}= \arg\min_{g} \big[\mathcal{L}_{s}(p(\mathbf{y}_s), q_s(\mathbf{y}_s)) + \mathcal{L}_{t}(p(\mathbf{y}_t))\big]
% \end{equation}
\begin{equation} \label{eqn:objective}
\small
\begin{split}
    \theta^{*} &= \arg\min_{\theta} \sum_{i,j} \big[\mathcal{L}_{s}(G(\mathbf{x}^i_s, \theta), \hat{\mathbf{y}}^i_s) + \mathcal{L}_{t}(G(\mathbf{x}^j_t, \theta))\big]\\
    &=\arg\min_{\theta} \Big[\mathbb{E}_{\mathbf{x}_s \sim p(\mathbf{x}_s), \hat{\mathbf{y}}_s \sim p(\hat{\mathbf{y}}_s)} \big[\mathcal{L}_{s}(G(\mathbf{x}_s, \theta), \hat{\mathbf{y}}_s)] \\
    &\quad \quad \quad \quad + \mathbb{E}_{\mathbf{x}_t \sim p(\mathbf{x}_t)} [\mathcal{L}_{t}(G(\mathbf{x}_t, \theta))\big]\Big]\\
    &=\arg\min_{\theta} \Big[\mathbb{E}_{\mathbf{y}_s \sim p(\mathbf{y}_s), \hat{\mathbf{y}}_s \sim p(\hat{\mathbf{y}}_s)} \big[\mathcal{L}_{s}(\mathbf{y}_s, \hat{\mathbf{y}}_s)] \\
    &\quad \quad \quad \quad + \mathbb{E}_{\mathbf{y}_t \sim p(\mathbf{y}_t)} [\mathcal{L}_{t}(\mathbf{y}_t)\big]\Big]\\
    % &=\arg\min_{\theta} \big[\mathcal{L}_{s}(\mathbf{y}_s, \hat{\mathbf{y}}_s) + \mathcal{L}_{t}(\mathbf{y}_t)\big]
\end{split}
\end{equation}
where $\theta$ is the parameters of $G$, $p(\mathbf{\cdot})$ is the probability density function. As the labels for $\mathcal{D}_s$ are available, $\mathcal{L}_s$ can be efficiently formulated as a supervised cross-entropy loss:
% as shown in Eqn. \eqref{Eqn2_crossentropy}.
%$\mathcal{L}_s$ denotes the supervised losses used for semantic segmentation and $\hat{\mathbf{y}}_s$ is the ground-truth of the corresponding input $\mathbf{x}_s$.
% where $\mathcal{L}_s$ is the supervised losses used for semantic segmentation and $q'_s(\mathbf{y}_s)$ \textcolor{red}{[where is q'?]} is the probability density function of real segmentation with the corresponding input $\mathbf{x}_{s}$, i.e. $\mathbf{y}_s = g(\mathbf{x}_s)$.
% The typical loss in this case is the cross entropy formulated as in Eqn. \eqref{Eqn2_crossentropy}.
% \begin{equation}
%     \small
%     \mathcal{L}_{s}(p(\mathbf{y}_s), q_s(\mathbf{y}_s)) = -{\mathbb{E}}_{\mathbf{x}_s \in \mathcal{X}_s} \left[\sum_{h,w,c} q'_s(\mathbf{y}^{h,w,c}_s)\log(p'(\mathbf{y}^{h,w,c}_s))\right]
% \end{equation}
\begin{equation}
    \small
    % \mathcal{L}_{s}(\mathbf{y}_s, \hat{\mathbf{y}}_s) = -{\mathbb{E}}_{\mathbf{x}_s \in \mathcal{X}_s} \left[\sum_{h,w,c} \hat{\mathbf{y}}^{h,w,c}_s\log\left(\mathbf{y}^{h,w,c}_s\right)\right]
    \mathcal{L}_{s}(\mathbf{y}_s, \hat{\mathbf{y}}_s) = -
    %\mathbb{E}_{\mathbf{y}_s \sim P(\mathbf{y}_s), \hat{\mathbf{y}}_s \sim P(\hat{\mathbf{y}}_s)} \left[
    \sum_{h,w, c} \hat{\mathbf{y}}^{h,w,c}_s\log\left(\mathbf{y}^{h,w,c}_s\right)
    %\right]
    \label{Eqn2_crossentropy}
\end{equation}
where $\mathbf{y}^{h,w,c}$ and $\hat{\mathbf{y}}^{h,w,c}$ represent the predicted and ground-truth probabilities of the pixel at the location of $(h, w)$  taking the label of $c$ , respectively. 
Meanwhile, $\mathcal{L}_t$ handles unlabeled data from
%case of unsupervised learning on 
the target domain where the ground-truth labels are not available. To alleviate this label lacking issue, several forms of $\mathcal{L}_{t}(\mathbf{y}_t)$ have been exploited such as cross-entropy loss with pseudo-labels \cite{zou2018unsupervised}, Probability Distribution Divergence (i.e. Adversarial loss defined via an additional Discriminator) \cite{tsai2018learning, tsai2019domain}, %\textcolor{red}{[CITE]}, 
or entropy formulation \cite{vu2019advent, pan2020unsupervised}.  

\noindent
\textbf{Entropy minimization revisited.}
%Some prior works \cite{vu2019advent, pan2020unsupervised} aim to minimize 
By adopting the \mbox{Shannon} entropy formulation to the target prediction and constraining function $G$ to produce a high-confident prediction,
%In this case, 
$\mathcal{L}_{t}$ can be formulated as %
% in Eqn. \eqref{eqn:entropy_loss}.
% \begin{equation} \label{eqn:entropy_loss}
%     \mathcal{L}_{t}(p(\mathbf{y}_t)) = {\mathbb{E}}_{\mathbf{x}_t \in \mathcal{X}_t}\left[\frac{-1}{\log(C)}\sum_{h,w,c}p'(\mathbf{y}^{h,w,c}_t)\log(p'(\mathbf{y}^{h,w,c}_t))\right]
% \end{equation}
\begin{equation} \label{eqn:entropy_loss}
\small
    % \mathcal{L}_{t}(\mathbf{y}_t) = 
    % %{\mathbb{E}}_{\mathbf{x}_t \in \mathcal{X}_t}\left[
    % % \frac{-1}{\log(C)}\sum_{h,w,c}\mathbf{y}^{h,w,c}_t\log\left(\mathbf{y}^{h,w,c}_t\right)
    % \frac{-1}{\log(C)}\sum_{h,w}\mathbf{y}^{h,w}_t\log\left(\mathbf{y}^{h,w}_t\right)
    % %\right]
    \mathcal{L}_{t}(\mathbf{y}_t) = 
    %{\mathbb{E}}_{\mathbf{x}_t \in \mathcal{X}_t}\left[
    % \frac{-1}{\log(C)}\sum_{h,w,c}\mathbf{y}^{h,w,c}_t\log\left(\mathbf{y}^{h,w,c}_t\right)
    \frac{-1}{\log(C)}\sum_{h,w,c}\mathbf{y}^{h,w,c}_t\log\left(\mathbf{y}^{h,w,c}_t\right).
    %\right]
\end{equation}
%
%
% The entropy loss leaves an issue here. Let's consider the derivative of $-\frac{\partial p'(\mathbf{y}^{h,w,c}_t)\log(p'(\mathbf{y}^{h,w,c}_t))}{\partial p'(\mathbf{y}^{h,w,c}_t)} = -\log(p'(\mathbf{y}^{h,w,c}_t)) - 1$. When $p'(\mathbf{y}^{h,w,c}_t) \to 0$, it is easy to recognize the derivative coming to $-\infty$. 
Although this form of $\mathcal{L}_{t}$ can give a direct assessment of the predicted segmentation maps, 
% this formulation 
it tends to be dominated 
% guided
by the high probability areas (since the high probability areas 
% have the 
produce a higher value 
% of
updated gradient due to  $\lim_{\mathbf{y}^{h,w,c}_t \to 1}\frac{-\partial \mathcal{L}_t(\mathbf{y}_t)}{\partial \mathbf{y}^{h,w,c}_t} = \frac{1}{\log(C)}$ and $\lim_{\mathbf{y}^{h,w,c}_t \to 0}\frac{-\partial \mathcal{L}_t(\mathbf{y}_t)}{\partial \mathbf{y}^{h,w,c}_t} = -\infty$),  %\textcolor{red}{[Dat: add supported info]},
% $\frac{\partial \mathcal{L}_t(\mathbf{y}_t)}{\partial \mathbf{y}^{h,w}_t} = -\log(\mathbf{y}^{h,w}_t) - 1$
i.e. easy classes, rather than difficult classes \cite{vu2019advent}. 
% More importantly, this is pixel-wised formulation where each pixel is independent to each other. 
More importantly, this is essentially a pixel-wise formation, where pixels are treated independently of each other.
% As a result, 
Consequently, the structural information is usually neglected in this form. This issue could lead to a confusion point during training process where two predicted segmentation maps have similar entropy but different segmentation accuracy, one correct and other incorrect as shown in Fig \ref{fig:same_entropy_1}.

\subsection{The Proposed UDS Metric}
%\hl{Image segmentation is pixel-level task so that using pixel-level loss makes sense and this has been proved by many pixel-lelvel network }

%\hl{I think you consider: using pixel-level requires alignment/corresponding between source and target}

%\hl{have you ever tried to combine both CE and BiMAL to take both local and lobal relations into consideration}

%\hl{BiMaL is more image distribution level}

%\hl{Also, be careful with the distribution assumption}

%$\to$ Modeling whole image instead of pixel independence.
% Rather than employing the pixel independent constraints to convert the image-level metric to pixel-level metric as entropy formulation, 
In the entropy formulation, the pixel independent constraints are employed to convert the image-level metric to pixel-level metric. In contrast,
we propose an image-level UDS metric that can directly evaluate the structural quality of $\mathbf{y}_t$. 
%during learning process, the whole structure of $\mathbf{y}_t$ can be directly modeled via its density function. 
Particularly, let $p_t(\mathbf{y}_t)$ and $q_t(\mathbf{y}_t)$ be the probability mass functions of the predicted distribution and the real (actual) distribution of the predicted segmentation map $\mathbf{y}_t$, respectively.
%
%In this section, we introduce our 
UDS metric measuring the efficiency of function $G$ on the target dataset can be expressed as follows:
%. The equation can be expressed as in Eqn. \eqref{eqn:UDSEqn}.
% \begin{equation} \label{eqn:DDS}
%     \operatorname{DDS} = \mathcal{H}(q_t) + \mathcal{D}_{KL}(q_t || p)
% \end{equation}
% \begin{equation}
%     \operatorname{DDS} = \int_{\mathcal{Y}} \mathcal{L}(q(\mathbf{y}), p(\mathbf{y}))q(\mathbf{y})d\mathbf{y} 
% \end{equation}
% \begin{equation}
%     \operatorname{DDS} = \int_{\mathcal{Y}_t} \mathcal{L}(\mathbf{y}_t, \hat{\mathbf{y}}_t)p_t(\mathbf{y}_t)d\mathbf{y}_t 
% \end{equation}
% \begin{equation}
%     \operatorname{DDS} = \int_{\mathcal{Y}_t} \mathcal{L}(\mathbf{y}_t, \hat{\mathbf{y}}_t)q_t(\hat{\mathbf{y}}_t)d\hat{\mathbf{y}}_t 
% \end{equation}
\begin{equation}
\begin{split}
    \operatorname{UDS}%(p_t(\mathbf{y}_t)) 
    &= \mathbb{E}_{\mathbf{y}_t \sim p(\mathbf{y}_t)} \mathcal{L_Y}\left(p_t(\mathbf{y}_t),q_t(\mathbf{y}_t)\right) \\
    &= \int \mathcal{L_Y}\left(p_t(\mathbf{y}_t),q_t(\mathbf{y}_t)\right) p_t(\mathbf{y}_t) d\mathbf{y}_t \;,
\end{split}
    \label{eqn:UDSEqn}
\end{equation}
where 
%$p_t(\mathbf{y}_t)$ and $q_t(\mathbf{y})$ denote the probability mass functions of the predicted distribution and the real (actual) distribution of the segmentation, 
$\mathcal{L_Y}\left(p_t(\mathbf{y}_t),q_t(\mathbf{y}_t)\right)$ defines the distance between %similarity of the
two distributions $p_t(\mathbf{y}_t)$ and $q_t(\mathbf{y}_t)$. Since there is no label for sample in the target domain, the direct access to $q_t(\mathbf{y}_t)$ is not available. 
% It should be noticed 
Note that although $\mathbf{x}_s$ and $\mathbf{x}_t$ could vary significantly in image space (e.g. difference in pixel appearance due to lighting, scenes, weather), their segmentation maps $\mathbf{y}_t$ and $\mathbf{y}_s$ share similar distributions in terms of both class distributions as well as global and local structural constraints (sky has to be above roads, trees should be on sidewalks, vehicles should be on roads, etc.). 
% as the labels of samples in target domain shares similar distribution as source domain, i.e. class distribution, image structure (
Therefore, one can practically adopted the prior knowledge learned from segmentation labels of the source domains for $q_t(\mathbf{y}_t)$ as
\begin{equation} \label{eqn:DDS_score_label_extend}
\begin{split}
    \operatorname{UDS} %(p_t(\mathbf{y}_t))
    % &= \int \mathcal{L_Y}\left(p_t(\mathbf{y}_t), q_t(\mathbf{y}_t)\right)dq_t(\mathbf{y}_t) \\
    &\approx \int \mathcal{L_Y}\left(p_t(\mathbf{y}_t),q_s(\mathbf{y}_t)\right) p_t(\mathbf{y}_t) d\mathbf{y}_t \;,
    %\int \mathcal{L_Y}\left(p_t(\mathbf{y}_t),q_s(\mathbf{y}_t)\right)dq_s(\mathbf{y}_t)
\end{split}
\end{equation}
% where $\left|\frac{f(\mathbf{y}_t)}{\partial \mathbf{y}_t}\right|$ is the Jacobian determinant of the function $\mathbf{f}(\mathbf{y}_t)$ with respect to $\mathbf{y}_t$, $p_z$ is the prior normal distribution having zero mean and unit variances,
% and $p(\mathbf{y}_t) = p_z(\mathbf{z}_t)\left|\frac{f(\mathbf{y}_t)}{\partial \mathbf{y}_t}\right|$ due to the change of variable theorem. $\mathcal{L_Z}$ is the distance metric used in the latent space. Learning the bijective $f$ will be further discussed in Sec. \ref{sec:bijective}.
% With respect to the usual Euclidean norm on the the Euclidean space, we adopt the $2$-Wasserstein distance between two distributes for $\mathcal{L_Z}$ that can be expressed as follows.
where the distribution $q_s(\mathbf{y}_t)$ 
is the probability mass functions of the real distribution learned from ground-truth segmentation maps of $\mathcal{D}_s$. %Section \ref{sec:bijective} 
%, that will be discussed in Section \ref{sec:bijective}. 
%As $q_s(\mathbf{y}_t)$ can be effectively learned from source domain (see Sec. \ref{sec:bijective}), 
As a result, the proposed USD metric can be computed without the requirement of labeled target data for learning the density of segmentation maps in target domain. 
There are several choices for $\mathcal{L_Y}$ to estimate the divergence between the two distributions $p_t(\mathbf{y}_t)$ and $q_s(\mathbf{y}_t)$. In this paper, we adopt the common metric such as  Kullback–Leibler (KL) formula for $\mathcal{L_Y}$. 
% Notice that other metrics are still applicable in the proposed UDS formulation.
Note that other metrics are also applicable in the proposed UDS formulation.
%
%In probabilistic theory, it is common to use Kullback–Leibler (KL) divergence function to measure the difference between two distributions. Hence, we adopt the distance $\mathcal{L_Y}$ as the KL divergence. 
% Moreover, in order to further improve 
Moreover, to enhance the smoothness of the predicted semantic segmentation, a regularization term $\tau$ is imposed into $\mathcal{L_Y}$ as
%In short, $\mathcal{L_Y}\left(p_t(\mathbf{y}_t),q_s(\mathbf{y}_t)\right)$ can be defined as in Eqn. \eqref{eqn:LYUpdated}.
%
\begin{equation} \label{eqn:LYUpdated}
\begin{split}
\small
    %\mathcal{L_Y}\left(p_t(\mathbf{y}_t),q_s(\mathbf{y}_t)\right) = \mathcal{D_{\text{KL}}}\left(p_t(\mathbf{y}_t) || q_s(\mathbf{y}_t)\right) + \tau(\mathbf{y}_t)
    \mathcal{L_Y}\left(p_t(\mathbf{y}_t),q_s(\mathbf{y}_t)\right) &= \log\left(\frac{p_t(\mathbf{y}_t)}{q_s(\mathbf{y}_t)}\right) + \tau(\mathbf{y}_t). \\
    % &= \log{p_t(\mathbf{y}_t)}) - \log({q_s(\mathbf{y}_t)}) + \tau(\mathbf{y}_t)
\end{split}
\end{equation}
%where $\mathcal{D_{KL}}$ is the KL divergence distance and $\tau$ is the regularization term. In Section \ref{sec:maximum_likelihood}, we will further discussed the regularization term. In addition, we will prove that our proposed Bijective Maximum Likelihood output is equivalent to a tight upper bound of the $\mathcal{L_Y}$.
By computing UDS, one can measure the quality of the predicted segmentation maps $\mathbf{y}_t$ on the target data. 

In the next sections, we firstly discuss in details the learning process of $q_s(\mathbf{y}_t)$, and then derivations of the UDS metric for the novel Bijective Maximum Likelihood loss. 

\subsection{Learning Distribution with Bijective Mapping on the Source Domain} \label{sec:bijective}

Let 
% $f: \mathbb{R}^{H \times W \times C} \to \mathbb{R}^{H \times W \times C}$ 
$F: \mathcal{Y} \to \mathcal{Z}$ 
be the bijective mapping function that maps a segmentation $\hat{\mathbf{y}}_s \in \mathcal{Y}$ to the latent space $\mathcal{Z}$, i.e. $\hat{\mathbf{z}}_s = F(\hat{\mathbf{y}}_s, \theta_F)$, where $\hat{\mathbf{z}}_s \sim q_z(\hat{\mathbf{z}}_s)$ is the latent variable, and $q_z$ is the prior distribution.
Then, the probability distribution $q_s(\hat{\mathbf{y}}_s)$ can be formulated via the change of variable formula:
% as in Eqn. \eqref{eqn:bijective_mapping}.
%
\begin{equation} \label{eqn:bijective_mapping}
\small
    \log(q_s(\hat{\mathbf{y}}_s)) = \log\left(q_{z}(\hat{\mathbf{z}}_s)\right) + \log\left(\left|\frac{\partial F(\hat{\mathbf{y}}_s, \theta_F)}{\partial \hat{\mathbf{y}}_s}\right|\right),
\end{equation}
where $\theta_F$ is the parameters of $F$, $\left|\frac{\partial F(\hat{\mathbf{y}}_s, \theta_F)}{\partial \hat{\mathbf{y}}_s}\right|$ denotes the Jacobian determinant of function $F(\hat{\mathbf{y}}_s, \theta_F)$ with respect to $\hat{\mathbf{y}}_s$. 
% follows the normal distribution $\mathcal{N}(\mathbf{0}, \mathbf{1})$. 
% In other to 
To learn the mapping function, the negative log-likelihood
% \cite{duong2016dam_cvpr, duong2019dam_ijcv, glow, truong2021fastflow}
will be minimized as follows:
\begin{equation} \label{eqn:BijectiveLearning}
\footnotesize
\begin{split}
    % f^{*} =& \arg\min_{f} {\mathbb{E}}_{\hat{\mathbf{y}}_s \in \mathcal{Y}_s} \Big[-\log(q_s(\mathbf{y}_s))\Big] \\
    % =& \arg\min_{f}{\mathbb{E}}_{\hat{\mathbf{y}}_s \in \mathcal{Y}_s} \left[-\log(q_{z}(\mathbf{z}_s)) - \log\left(\left|\frac{\partial f(\mathbf{y}_s)}{\partial \mathbf{y}_s}\right|\right)\right]
    \theta_F^{*} =& \arg\min_{\theta_F} %{\mathbb{E}}_{\hat{\mathbf{y}}_s \in \mathcal{Y}} \Big[-\log(q_s(\hat{\mathbf{y}}_s))\Big] \\
    \mathbb{E}_{\hat{\mathbf{y}}_s \sim q_s(\hat{\mathbf{y}}_s)} \Big[-\log(q_s(\hat{\mathbf{y}}_s))\Big] \\
    =& \arg\min_{\theta_F} \mathbb{E}_{\hat{\mathbf{z}}_s \sim q_z(\hat{\mathbf{z}}_s)} \left[-\log\left(q_{z}(\hat{\mathbf{z}}_s)\right) - \log\left(\left|\frac{\partial F(\hat{\mathbf{y}}_s, \theta_F)}{\partial \hat{\mathbf{y}}_s}\right|\right)\right].
    \raisetag{40pt}
\end{split}
\end{equation}
% Then, the probability distribution $p_z(\mathbf{y}_z)$ can be formulated via the change of variable formula as follows.
% \begin{equation} \label{eqn:bijective_mapping}
%     \log(p(\mathbf{y}_s)) = \log\pi_{z}(\mathbf{z}_s) + \log\left(\left|\frac{\partial f(\mathbf{y}_s)}{\partial \mathbf{y}_s}\right|\right)
% \end{equation}
% where $\left|\frac{\partial f(\mathbf{y}_s)}{\partial \mathbf{y}_s}\right|$ is the Jacobian determinant of the the function $f(\mathbf{y}_s)$ with respect to $\mathbf{y}_s$. 
In general, there are various choices for the prior distribution $q_z$. However, the ideal distribution should satisfy two 
% following properties:
criteria: (1) simplicity in the density estimation, and (2) easy in sampling. 
% Taking these properties into consideration, 
Considering the two criteria, we choose Normal distribution %\cite{duong2016dam_cvpr, duong2019dam_ijcv, Duong_2017_ICCV, 9108692} 
as the prior distribution $q_z$. 
% It is noticed that any other distribution can be chosen as long as it satisfies the mentioned properties.
Note that any other distribution is also feasible as long as it satisfies the mentioned criteria.

% The prior distributions pz. In general, there are various
% choices for the prior distribution pz and the ideal one should
% have two properties: (1) simplicity in density estimation,
% and (2) easily sampling. Motivated from these properties,
% we choose Gaussian distribution for pz. Notice that other
% distribution types are still applicable in our framework.

To enforce the information flow from a segmentation domain to a latent space with different abstraction levels, the bijective function $F$ can be further formulated as a composition of several sub-bijective functions $f_i$ as $F = f_1 \circ f_2 \circ ... \circ f_K$, 
% \begin{equation} \label{eqn:MappingFunction}
% \small
%     f = f_1 \circ f_2 \circ ... \circ f_N
% \end{equation}
where $K$ is the number of sub-functions. The Jacobian $\frac{\partial F}{\partial \mathbf{y}_s}$ can be derived by $\frac{\partial F}{\partial \hat{\mathbf{y}}_s} = \frac{\partial f_1}{\partial \hat{\mathbf{y}}_s} \cdot \frac{\partial f_2}{ \partial f_1} \cdots \frac{\partial f_K}{ \partial f_{K-1}}$. With this structure, the properties of each $f_i$ will define the properties for the whole bijective mapping function $F$. 
% For example, if the Jacobian of $\frac{\partial f_1}{\partial \mathbf{x}}$ is tractable, then $\mathcal{F}$ is also tractable.
Interestingly, with this form, $F$ becomes a DNN structure when $f_i$ is a non-linear function built from a composition of convolutional layers. Several DNN structures \cite{dinh2015nice, dinh2017density,Duong_2017_ICCV, duong2020vec2face, glow, duong2019learning, truong2021fastflow} can be adopted for sub-functions.

\section{Bijective Maximum Likelihood Loss} %(BiMAL) Loss} \label{sec:maximum_likelihood}

In this section, we present the proposed Bijective Maximum Likelihood (BiMaL) which can be used as the loss of target domain $\mathcal{L}_t$. 
%Now, Let's consider the form of KL divergence in 
From Eqns. \eqref{eqn:DDS_score_label_extend} and \eqref{eqn:LYUpdated}, %the first term of 
UDS metric can be rewritten as follows: %the KL-divergence can be rewritten as follows.
\begin{equation} \label{eqn:KL_to_llk}
\small
\begin{split}
        %\mathcal{D}_{KL}&\left(p_t(\mathbf{y}_t) || q_s(\mathbf{y}_t)\right) = 
        \text{UDS} &= \int \left[\log\left(\frac{p_t(\mathbf{y}_t)}{q_s(\mathbf{y}_t)}\right)+  \tau(\mathbf{y}_t)\right]p_t(\mathbf{y}_t) d\mathbf{y}_t \\
        & = {\mathbb{E}}_{\mathbf{y}_t \sim p_t(\mathbf{y}_t)}\left[
        \log(p_t(\mathbf{y}_t))\right] \\
        & \quad - {\mathbb{E}}_{\mathbf{y}_t \sim p_t(\mathbf{y}_t)} \left[\log(q_s(\mathbf{y}_t)) \right] 
        +  {\mathbb{E}}_{\mathbf{y}_t \sim p_t(\mathbf{y}_t)}\left[\tau(\mathbf{y}_t)\right] \\
        % &= {\mathbb{E}}_{\mathbf{y}_t \sim p_t(\mathbf{y}_t)}\left[
        % \log(p_t(\mathbf{y}_t)) - \log(q_s(\mathbf{y}_t)) +  \tau(\mathbf{y}_t)\right] \\
        % &= {\mathbb{E}}_{\mathbf{z}_t \sim p_z(\mathbf{z}_t)}\left[
        % \log(p_z(\mathbf{z}_t)) - \log(q_z(\mathbf{z}_t)) +  \tau(\mathbf{y}_t)\right]\\
        % &= {\mathbb{E}}_{\mathbf{y}_t \sim p_t(\mathbf{y}_t)}\left[
        % \log(p_t(\mathbf{y}_t) )\right] - {\mathbb{E}}_{\mathbf{z}_t \sim p_z(\mathbf{z}_t)} \left[\log(q_z(\mathbf{z}_t))\right]\\
        % &< - {\mathbb{E}}_{\mathbf{z}_t \sim p_z(\mathbf{z}_t)} \left[\log(q_z(\mathbf{z}_t))\right]  
        &\leq {\mathbb{E}}_{\mathbf{y}_t \sim p_t(\mathbf{y}_t)} \left[-\log(q_s(\mathbf{y}_t)) + \tau(\mathbf{y}_t)\right]  
        % +  {\mathbb{E}}_{\mathbf{y}_t \sim p_t(\mathbf{y}_t)}\left[\tau(\mathbf{y}_t)\right] \\
        % & {\mathbb{E}}_{\mathbf{z}_t \sim p_z(\mathbf{z}_t)}\left[
        %  \log(q_z(\mathbf{z}_t))\right] ???
\end{split}    
\end{equation}

It should be noticed that with any form of the distribution $p_t$, the above inequality still holds as $p_t(\mathbf{y}_t) \in [0,1]$ and $\log(p_t(\mathbf{y}_t)) \leq 0$. Now, we define our Bijective Maximum Likelihood Loss as
\begin{equation} \label{eqn:BiMaL}
\begin{split}
    \mathcal{L}_t(\mathbf{y}_t) = -\log(q_s(\mathbf{y}_t))+ \tau(\mathbf{y}_t),
    %\mathcal{L}_{llk}(\mathbf{y}_t) + \tau(\mathbf{y}_t) \\
\end{split}
\end{equation}
% Thus, the second term of Eqn. \eqref{eqn:objective} becomes.
where $\log(q_s(\mathbf{y}_t))$ defines the log-likelihood of $\mathbf{y}_t$ with respect to the density function $q_s$.
Then, by adopting the bijectve function $F$ learned from Eqn. \eqref{eqn:BijectiveLearning} using samples from source domain and the prior distribution $q_z$, the first term of $\mathcal{L}_t(\mathbf{y}_t)$ in Eqn. \eqref{eqn:BiMaL} can be efficiently computed via log-likelihood formulation:
\begin{equation} \label{eqn:define_llk}
\small
\begin{split}
    % \mathcal{L}_{llk}(\mathbf{y}_t) &= {\mathbb{E}}_{\mathbf{y}_t \sim p_t(\mathbf{y}_t)} \left[-\log(q_s(\mathbf{y}_t))\right] \\
    % %- {\mathbb{E}}_{\mathbf{z}_t \sim p_z(\mathbf{z}_t)} \left[\log(q_z(\mathbf{z}_t))\right] \\
    % &= {\mathbb{E}}_{\mathbf{z}_t \sim p_z(\mathbf{z}_t)} \left[ -\log\left(q_{z}(\mathbf{z}_t)\right) - \log\left(\left|\frac{\partial F(\mathbf{y}_t, \theta_F)}{\partial \mathbf{y}_t}\right|\right) \right]
    \mathcal{L}_{llk}(\mathbf{y}_t) &= -\log(q_s(\mathbf{y}_t)) \\
    %- {\mathbb{E}}_{\mathbf{z}_t \sim p_z(\mathbf{z}_t)} \left[\log(q_z(\mathbf{z}_t))\right] \\
    &=  -\log\left(q_{z}(\mathbf{z}_t)\right) - \log\left(\left|\frac{\partial F(\mathbf{y}_t, \theta_F)}{\partial \mathbf{y}_t}\right|\right), 
\end{split}
\end{equation}
where $\mathbf{z}_t = F(\mathbf{y}_t, \theta_F)$. Thanks to the bijective property of the mapping function $F$, the minimum negative log-likelihood loss $\mathcal{L}_{llk}(\mathbf{y}_t)$
%, instead of directly computing negative log-likelihood on the segmentation domain where we do not the form of $q_s$, 
can be effectively computed via 
%the latent space where the 
the density of the prior distribution $q_z$ and its associated Jacobian determinant $\left|\frac{\partial F(\mathbf{y}_t, \theta_F)}{\partial \mathbf{y}_t}\right|$. For the second term of $\mathcal{L}_t(\mathbf{y}_t)$, we further enhance the smoothness of the predicted semantic segmentation with the pair-wised formulation to encourage similar predictions for neighbourhood pixels with similar color:
%is explicitly defined and 
% Hence, we adopt the bijective maximum likelihood as the loss used for $\mathcal{L}_t$ defined as follows.
%
% where 
\begin{equation} \label{eqn:define_tau}
\scriptsize
    \tau(\mathbf{y}_t) = \sum_{h,w}\sum_{h', w'} \exp \left\{-\frac{||\mathbf{x}_t^{h,w} -  \mathbf{x}_t^{h', w'}||_2^2}{2\sigma_1^2} - \frac{||\mathbf{y}_t^{h,w} - \mathbf{y}_t^{h', w'}||_2^2}{2\sigma_2^2}\right\}%||\mathbf{y}_t^{h,w} - \mathbf{y}_t^{h', w'}||_2^2
\end{equation}
where $(h', w') \in \mathcal{N}_{h, w}$ denotes the neighbourhood pixels of $(h, w)$, $\mathbf{x}^{h,w}$ represents the color at pixel $(h, w)$; and $\{\sigma_1, \sigma_2\}$ are the hyper parameters controlling the scale of Gaussian kernels. 
It should be noted that any regularizers \cite{chen2018deeplab, duong2029cvpr_automatic} enhancing the smoothness of the segmentation results can also be adopted for $\tau$.   
%The regularization term $\tau$ here impose the smoothness of the segmentation such that the close pixels having similar color should have the similar predictions.
%
% As shown in Eqn \eqref{eqn:KL_to_llk}, our BiMaL $\mathcal{L}_{llk}$ is a tight upper bound of $\mathcal{D}_{KL}&\left(p_t(\mathbf{y}_t) || q_s(\mathbf{y}_t)\right)$. Equivalently, our loss $\mathcal{L}_t$ is a tight upper bound of $\mathcal{L_Y}$. %
% In another word, if we minimize our proposed loss, it will also minimize the loss $\mathcal{L_Y}$. %Unaligned Domain Score.
Putting Eqns. \eqref{eqn:BiMaL}, \eqref{eqn:define_llk}, \eqref{eqn:define_tau} to Eqn \eqref{eqn:objective}, the objective function can be rewritten as: % follows:
\begin{equation}
\small
\begin{split}
    % g^{*}= \arg\min_{g} \big[\mathcal{L}_{s}(\mathbf{y}_s, \hat{\mathbf{y}}_s) + \mathcal{L}_{llk}(\mathbf{y}_t) + \tau(\mathbf{y}_t)\big]
        % \theta^* = \arg\min_{\theta}\sum_{i, j}\big[ &\mathcal{L}_{s}(G(\mathbf{x}^i_s, \theta), \hat{\mathbf{y}}^i_s) + \mathcal{L}_{llk}(G(\mathbf{x}^j_t, \theta)) \\
        % &+ \tau(G(\mathbf{x}^j_t, \theta))\big] \\
         \theta^*=\arg\min_{\theta} \Big[&\mathbb{E}_{\mathbf{y}_s \sim p(\mathbf{y}_s), \hat{\mathbf{y}}_s \sim p(\hat{\mathbf{y}}_s)} \big[\mathcal{L}_{s}(\mathbf{y}_s, \hat{\mathbf{y}}_s)] \\ &+ \mathbb{E}_{\mathbf{y}_t \sim p(\mathbf{y}_t)} [\mathcal{L}_{llk}(\mathbf{y}_t) + \tau(\mathbf{y}_t)\big]\Big]
\end{split}
\end{equation}
Figure \ref{fig:proposed_framework} illustrates our proposed BiMaL framework to learn the deep segmentation network $G$. Also, we can prove that direct entropy minimization as Eqn. \eqref{eqn:entropy_loss} is just a particular case of our log likelihood maximization. We will further discuss how our maximum likelihood can cover the case of pixel-independent entropy minimization in Section \ref{sec:MLE_Entropy}.

\noindent
\subsection{BiMaL properties}  \label{sec:MLE_Entropy}
%
% \newline

% \noindent
\textit{\textbf{Global Structure Learning.}} Sharing similar property with \cite{duong2016dam_cvpr, duong2019dam_ijcv, duong2020vec2face, Duong_2017_ICCV, 9108692}, from Eqn. \eqref{eqn:bijective_mapping}, as the learned density function is adopted for the entire segmentation map $\hat{\mathbf{y}}_s$, the global structure in $\hat{\mathbf{y}}_s$ can be efficiently captured and modeled.

%Thanks to the designed bijection F, the complex distribution of segmentation of segmentation maps can be efficiently captured. Moreover, as the density function is adopted for the whole segmentation map $\hat{\mathbf{y}}_s$, the Bijective function can capture

% our proposed graphical model has the capability of modeling arbitrary complex data distributions while keeping the inference process tractable. Fur- thermore, from Eqns. (6) and (9), the mapping function is invertible. Therefore, both inference (i.e. mapping from im- age to latent space) and generation (i.e. from latent to image space) are exact and efficient.

% \noindent
\textit{\textbf{Tractability and Invertibility.}} Thanks to the designed bijection F, the complex distribution of segmentation maps can be efficiently captured. Moreover, the mapping function is bijective, and, therefore, both inference and generation  are exact and tractable.
\subsection{Relation to Entropy Minimization} \label{sec:MLE_Entropy}
The first term of UDS in Eqn. \eqref{eqn:KL_to_llk} can be derived as
% In this section, we will show how our proposed BiMaL loss can be a generalized form of the direct entropy minimization. Initially, let us revise the formula of entropy minimization defined as follows.
% \begin{equation} \label{eqn:entropy_min}
% \small
% \begin{split}
%     \operatorname{Entropy} &= -\sum_{\mathbf{y}_t \sim p_t(\mathbf{y}_t)} p_t(\mathbf{y}_t)\log(p_t(\mathbf{y}_t)) \\
%     &= - {\mathbb{E}}_{\mathbf{y}_t \sim p_t(\mathbf{y}_t)} \log(p_t(\mathbf{y}_t) \\
% \end{split} 
% \end{equation}
% If we take the assumption on pixel independence, Eqn \eqref{eqn:entropy_min} can be further extended as Eqn \eqref{eqn:entropy_loss}. Also, we always know that $\mathcal{D_{\text{KL}}} \geq 0$, 
\begin{equation}
\small
    \begin{split}
        % \mathcal{D_KL} &\geq 0 \\
        %\Leftrightarrow
        & \int \log\left(\frac{p_t(\mathbf{y})}{q_s(\mathbf{y})}\right)p_t(\mathbf{y}_t) d\mathbf{y}_t \geq 0 \\
        \Leftrightarrow & {\mathbb{E}}_{\mathbf{y}_t \sim p_t(\mathbf{y}_t)}\left[\log(p_t(\mathbf{y}_t)) - \log(q_s(\mathbf{y}_t))\right] \geq 0 \\
        % {\mathbb{E}}_{\mathbf{y}_t \sim p_t(\mathbf{y}_t)}\left[\log(p_t(\mathbf{y}_t)) - \log(q_s(\mathbf{y}_t))\right] \\
        \Leftrightarrow & {\mathbb{E}}_{\mathbf{y}_t \sim p_t(\mathbf{y}_t)}\left[-\log(q_s(\mathbf{y}_t))\right] \geq {\mathbb{E}}_{\mathbf{y}_t \sim p_t(\mathbf{y}_t)}\left[-\log(p_t(\mathbf{y}_t))\right]\\
        \Leftrightarrow & \mathbb{E}_{\mathbf{y}_t \sim p_t(\mathbf{y}_t)}[\mathcal{L}_{llk}(\mathbf{y}_t)] \geq \text{Ent}(\mathbf{Y}_t)%\operatorname{Entropy}
        % &= {\mathbb{E}}_{\mathbf{z}_t \sim p_z(\mathbf{z}_t)}\left[
        % \log(p_z(\mathbf{z}_t)) - \log(q_z(\mathbf{z}_t))\right]\\
        % &= {\mathbb{E}}_{\mathbf{y}_t \sim p_t(\mathbf{y}_t)}\left[
        % \log(p_t(\mathbf{y}_t) )\right] - {\mathbb{E}}_{\mathbf{z}_t \sim p_z(\mathbf{z}_t)} \left[\log(q_z(\mathbf{z}_t))\right]\\
        % &\leq - {\mathbb{E}}_{\mathbf{z}_t \sim p_z(\mathbf{z}_t)} \left[\log(q_z(\mathbf{z}_t))\right]
        \raisetag{40pt}
    \end{split}
\end{equation}
where $\mathbf{Y}_t$ is the random variable with possible values $\mathbf{y}_t \sim p_t(\mathbf{y}_t)$, and $\text{Ent}(\mathbf{Y}_t)$ denotes the entropy of the random variable $\mathbf{Y}_t$.
It can be seen that the proposed negative log-likelihood $\mathcal{L}_{llk}$ is an upper bound of the entropy of $\mathbf{Y}_t$. Therefore, minimizing our proposed BiMaL loss will also enforce the entropy minimization process. 
% Moreover, without the assumption of pixel independence, our proposed BiMaL has more capabilities of modeling and evaluating structural information in image-level than previous pixel-level approaches \cite{vu2019advent, pan2020unsupervised, chen2019domain}.
%Moreover, without the assumption of pixel independence, our proposed BiMaL can model and evaluate structural information at the image-level better than previous pixel-level approaches \cite{vu2019advent, pan2020unsupervised, chen2019domain}.
Moreover, by not assuming pixel independence, our proposed BiMaL can model and evaluate structural information at the image-level better than previous pixel-level approaches \cite{chen2019domain, pan2020unsupervised, vu2019advent}.

\section{Experimental Results}

This section will present our experimental results on three different benchmarks, i.e. SYNTHIA $to$ Cityscapes, GTA $to$ Cityscapes, and SYNTHIA $to$ Vistas. First, we overview datasets and network architectures used in our experiments. Second, we present the ablation study to analyze the effectiveness of our proposed BiMaL and the capability of the bijective network. Finally, we present the quantitative and qualitative results of our method compared to prior methods on the three benchmarks.

\begin{figure}[t]
    \centering
    \includegraphics[width=0.45\textwidth]{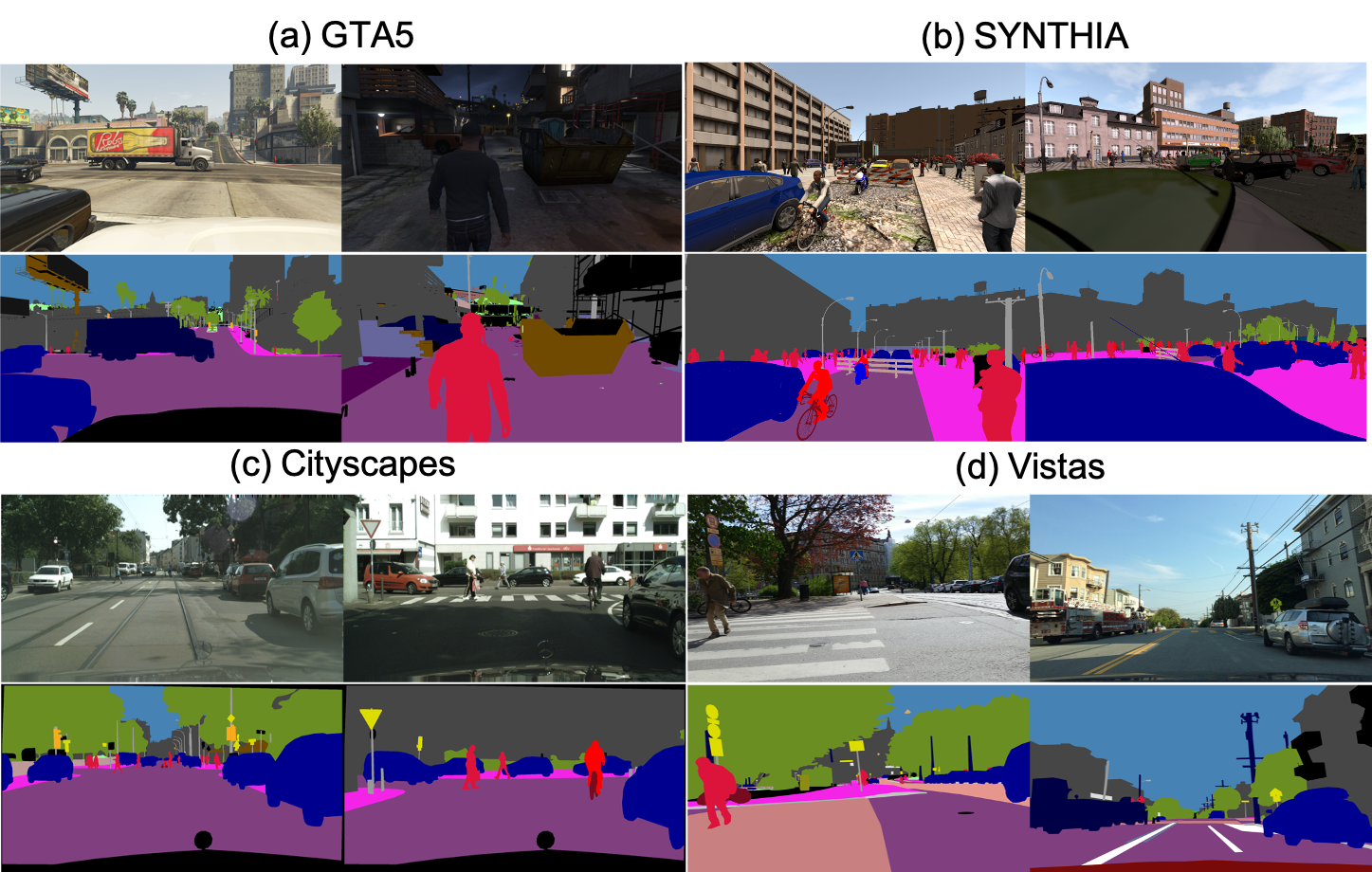}
    \caption{Examples of four semantic segmentation datasets including RGB images (top row) and corresponding ground-truth images (bottom row). (a) GTA and (b) SYNTHIA are synthetic datasets, (c) Cityscapes and (d) Vistas are real collected datasets.} % \hl{this image can be removed}}
    \label{fig:database}
    \vspace{-4mm}
\end{figure}

\subsection{Datasets}

\textbf{GTA5} \cite{Richter_2016_ECCV} is a synthetic dataset containing $24,966$ densely labelled images at the resolution of $1,914 \times 1,052$. This dataset was collected from the game Grand Theft Auto V. The ground-truth annotations were automatically generated with 33 categories. In our experiments, we consider 19 categories that are compatible with the Cityscapes \cite{cordts2016cityscapes}.

% GTA5:  The synthetic dataset GTA5 [19] contains
% 24,966 synthetic images with a resolution of 1914 ×
% 1052px and corresponding ground-truth annotations.
% These synthetic images are collected from a video
% game based on the urban scenery of Los Angeles
% city. The ground-truth annotations generated automatically contain 33 categories. For training, we consider only 19 categories which are compatible with the
% Cityscapes dataset [6], similarly to previous work.

\textbf{SYNTHIA (SYNHIA-RAND-CITYSCAPES) }\cite{Ros_2016_CVPR} is also synthetic dataset that contains $9,400$ pixel-level labelled RGB images. In our experiments, we use the 16 common categories that overlap with the Cityscapes dataset.

\textbf{Cityscapes} \cite{cordts2016cityscapes} is a real-world dataset including $3,975$ images with fine semantic, dense pixel annotations of 30 classes.
In our experiments, $2,495$ images are used for training and $500$ images are used for testing.

% \textcolor{red}{This text was copied from another paper}
% SYNTHIA: SYNTHIA-RAND-CITYSCAPES [20] is
% used as another synthetic dataset. It contains 9,400
% fully annotated RGB images. During the training
% time, we consider the 16 common categories with the Cityscapes dataset. During evaluation, 16- and 13-class subsets are used to evaluate the performance.

% Cityscapes: As the dataset collected from real world,
% Cityscapes [6] provides 3975 images with fine segmentation annotations. 2975 images are taken from
% the training set of Cityscapes to be used for training.
% The 500 images from the evaluation set of Cityscapes
% are used to evaluate the performance of our model.

\textbf{Vistas (Mapillary Vistas Dataset)} \cite{MVD2017} is diverse street-level imagery dataset with pixel‑accurate and instance‑specific human annotations for understanding street scenes around the world. Vistas consists of $25,000$ high-resolution images and $124$ semantic object categories. In our experiments, we consider 7 classes that are common to SYNTHIA, Cityscapes and Vistas as shown in Fig. \ref{fig:database}.

\begin{figure}[t]
    \centering
    \includegraphics[width=0.48\textwidth]{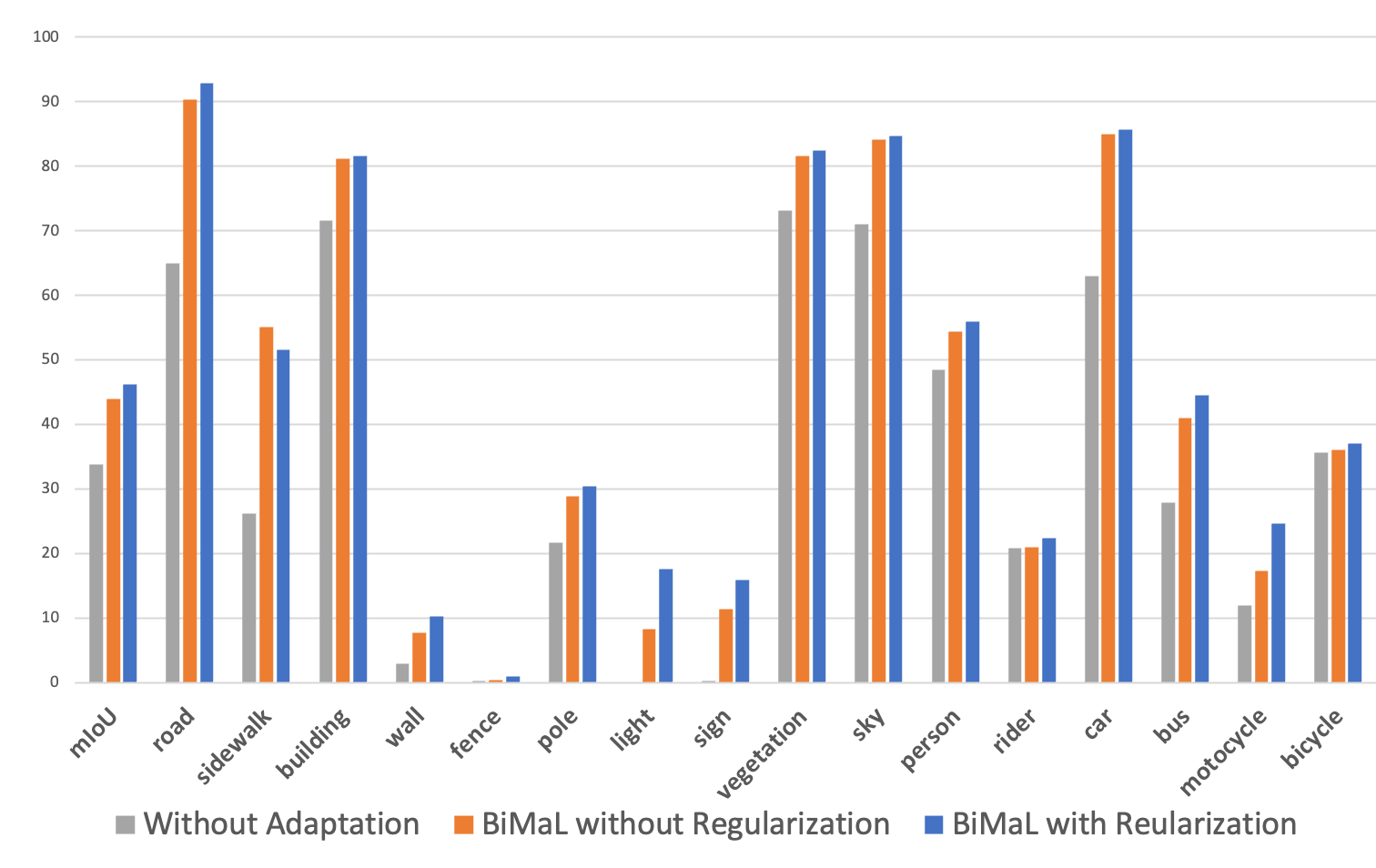}
    \caption{Ablative semantic segmentation performance mIoU (\%) on the effectiveness of the proposed BiMaL loss.}
    % Semantic segmentation performance mIoU (\%) on Cityscapes validation set of models trained on SYNTHIA}
    \label{fig:ablation_loss}
\end{figure}

\begin{figure}[t]
    \centering
    \includegraphics[width=0.45\textwidth]{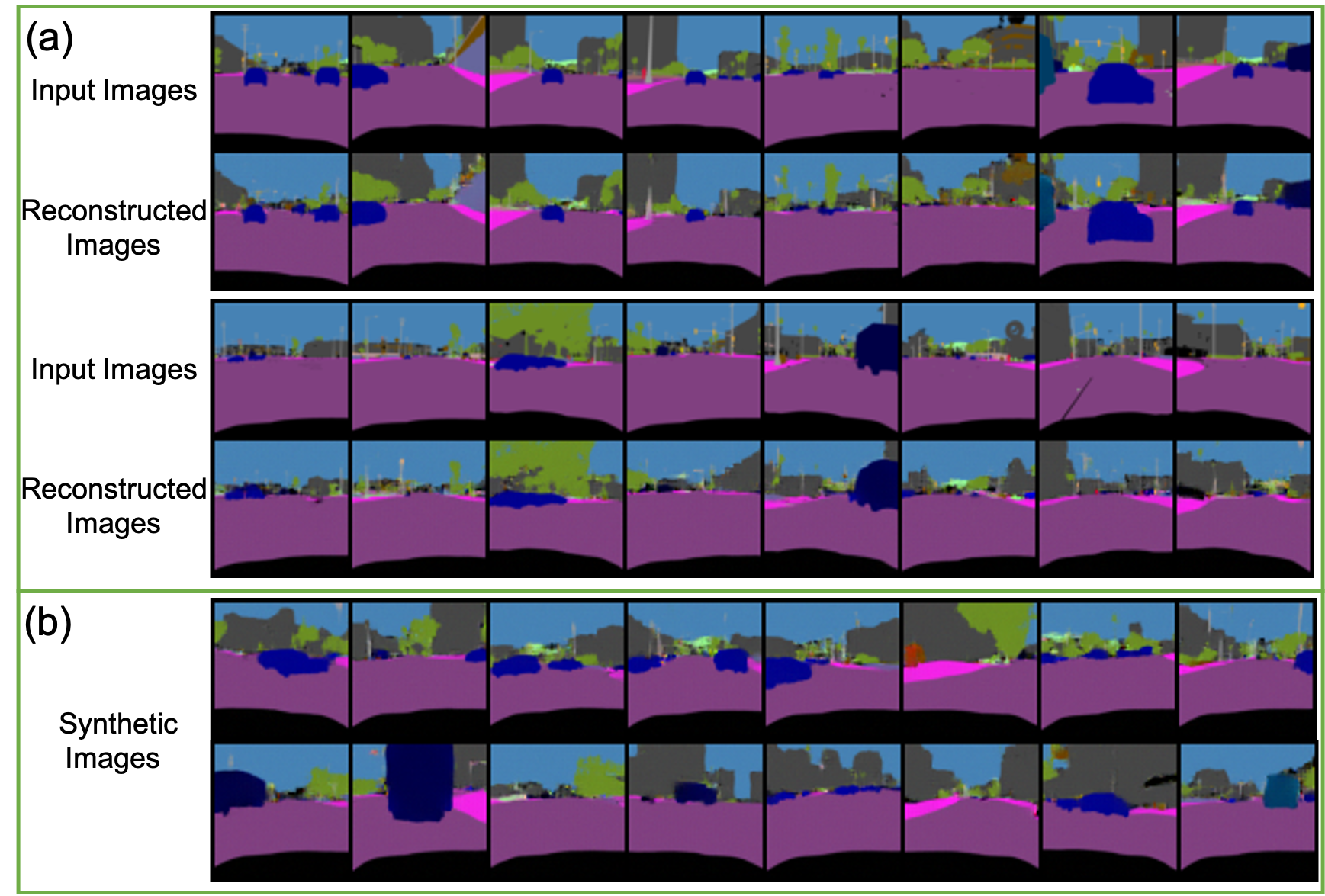}
    \caption{\textbf{Reconstructed Images and Synthetic Images From The Bijective Mapping Function $F$}. (a) Reconstructed images (bottom row) from the corresponding input images (top row). (b) Synthetic images sampled from the latent space.} % \hl{why the image looks blurry}} %\textcolor{red}{[This Fig has not ref in text yet.]}}
    \label{fig:rec_images}
    \vspace{-4mm}
\end{figure}

\begin{table*}[t]
        % \scriptsize
        \centering
        \caption{\textbf{Semantic segmentation performance mIoU (\%) on Cityscapes validation set of different models trained on SYNTHIA}. We also show the mIoU (\%) of the $13$ classes (mIoU*) excluding classes with *.}
            \resizebox{\textwidth}{!}{%
            \begin{tabular}{|c|c|c|c|c|c|c|c|c|c|c|c|c|c|c|c|c|c|c|c|c|c|c|c|}
				\multicolumn{19}{c}{ SYNTHIA $\rightarrow$ Cityscapes (16 classes)}\\
				\hline
				Models  & \rotatebox{90}{\textbf{road}} & \rotatebox{90}{\textbf{sidewalk}} & \rotatebox{90}{\textbf{building}} & \rotatebox{90}{\textbf{wall*}} & \rotatebox{90}{\textbf{fence*}} & \rotatebox{90}{\textbf{pole*}} & \rotatebox{90}{\textbf{light}} & \rotatebox{90}{\textbf{sign}} & \rotatebox{90}{\textbf{veg}} & \rotatebox{90}{\textbf{sky}} & \rotatebox{90}{\textbf{person}} & \rotatebox{90}{\textbf{rider}} & \rotatebox{90}{\textbf{car}} & \rotatebox{90}{\textbf{bus}} & \rotatebox{90}{\textbf{mbike}} & \rotatebox{90}{\textbf{bike}} & \rotatebox{90}{\textbf{mIoU}} & \rotatebox{90}{\textbf{mIoU*}}\\
				\hline
				Without Adaptation & 64.9 & 26.1 & 71.5 & 3.0 & 0.2 & 21.7 & 0.1 & 0.2 & 73.1 & 71.0 & 48.4 & 20.7 & 62.9 & 27.9 & 12.0 & 35.6 & 33.7 & 39.6 \\
				SPIGAN-no-PI~\cite{lee2018spigan}&69.5&29.4&68.7&4.4&0.3&32.4&5.8&15.0&81.0&78.7&52.2&13.1&72.8&23.6&7.9&18.7&35.8&41.2\\
				SPIGAN~\cite{lee2018spigan}&71.1&29.8&71.4&3.7&0.3&\textbf{33.2}&6.4&{15.6}&81.2&78.9&52.7&13.1&75.9&25.5&10.0&20.5&36.8&42.4\\
				% \hline
				AdaptSegnet~\cite{tsai2018learning}&79.2&37.2&78.8&-&-&-&9.9&10.5&78.2&80.5&53.5&19.6&67.0&29.5&21.6&31.3&-&45.9\\
				AdaptPatch~\cite{tsai2019domain}&82.2&39.4&79.4&-&-&-&6.5&10.8&77.8&82.0&54.9&21.1&67.7&30.7&17.8&32.2&-&46.3\\
				CLAN~\cite{luo2018taking}&81.3&37.0&80.1&-&-&-&{16.1}&13.7&78.2&81.5&53.4&21.2&73.0&32.9&{22.6}&30.7&-&47.8\\
				AdvEnt~\cite{vu2019advent}&87.0&44.1&79.7&{9.6}&{0.6}&24.3&4.8&7.2&80.1&83.6&{56.4}&\textbf{23.7}&72.7&32.6&12.8&33.7&40.8&47.6\\
				% \rowcolor[gray]{.92}[0pt][0pt]\rule{0pt}{3ex}
				IntraDA \cite{pan2020unsupervised} & 84.3 & 37.7 & 79.5 & 5.3 & 0.4 & 24.9 & 9.2 & 8.4 & 80.0 & 84.1 & \textbf{57.2} & 23.0 & 78.0 & 38.1 & 20.3 & 36.5 & 41.7 & 48.9 \\
				DADA\cite{vu2019dada} &{89.2}&{44.8}&{81.4}&6.8&0.3&26.2&8.6&11.1&{81.8}&{84.0}&54.7&19.3&{79.7}&{40.7}&14.0&{38.8}&{42.6}&{49.8} \\
				% \textbf{Ours} &\textbf{90.3}&\textbf{49.0}&\textbf{81.5}&8.5&0.3&{30.3}&10.6&11.5&\textbf{82.3}&\textbf{84.3}&55.7&18.2&\textbf{85.0}&\textbf{44.5}&21.3&31.0&\textbf{44.0}&\textbf{51.2} \\
				% \textbf{Our BiMaL} & \textbf{91.5} & \textbf{51.5} & \textbf{81.5} &  7.6 & 0.2 & 29.8 & 10.0 &  10.3 & \textbf{82.4} & \textbf{84.6} & 55.9 & 18.6 & \textbf{85.7} & \textbf{44.5} &  20.6 & 30.8 & \textbf{44.1} & \textbf{51.3}\\
				% CCM \cite{} & 79.6 & 36.4 & 80.6 & 13.3 & 0.3 & 25.5 & 22.4 & 14.9 & 81.8 & 77.4 & 56.8 & 25.9 & 80.7 & 45.3 & 29.9 & 52.0 & 45.2 & 52.9 \\
				% CAG-UDA \cite{zhang2019category} & 84.7 & 40.8 & 81.7 & 7.8 & 0.0 & 35.1 & 13.3 & 22.7 & 84.5 & 77.6 & 64.2 & 27.8 & 80.9 & 19.7 & 22.7 & 48.3 & 44.5 & 51.5 \\ 
				% CCM \cite{li2020content} & 79.6 & 36.4 & 80.6 & 13.3 & 0.3 & 25.5 & 22.4 & 14.9 & 81.8 & 77.4 & 56.8 & 25.9 & 80.7 & 45.3 & 29.9 & 52.0 & 45.2 & 52.9 \\
				\textbf{Our BiMaL} & \textbf{92.8} & \textbf{51.5} & \textbf{81.5} & \textbf{10.2} & \textbf{1.0} & 30.4 & \textbf{17.6} & \textbf{15.9} & \textbf{82.4} & \textbf{84.6} & 55.9 & 22.3 & \textbf{85.7} & \textbf{44.5} & \textbf{24.6} & \textbf{38.8} & \textbf{46.2} & \textbf{53.7} \\
				\hline
			\end{tabular}
			}
			\label{tab:synthia2city} 
\end{table*}

\begin{table*}[t]
    \centering
     \caption{\textbf{Semantic segmentation performance mIoU (\%) on Cityscapes validation set of different models trained on GTA5}}
    \resizebox{\textwidth}{!}{
    \begin{tabular}{|c|c|c|c|c|c|c|c|c|c|c|c|c|c|c|c|c|c|c|c|c|}
        \multicolumn{21}{c}{GTA5 $\to$ Cityscapes (19 classes)}\\
        % \hline \hline
        % \toprule[1.0pt]
        \hline 
        Models & \rotatebox{90}{\textbf{road}} & \rotatebox{90}{\textbf{sidewalk}} & \rotatebox{90}{\textbf{building}} & \rotatebox{90}{\textbf{wall}} & \rotatebox{90}{\textbf{fence}} & \rotatebox{90}{\textbf{pole}} & \rotatebox{90}{\textbf{light}} & \rotatebox{90}{\textbf{sign}} & \rotatebox{90}{\textbf{\textbf{veg}}} & \rotatebox{90}{\textbf{terrain}} & \rotatebox{90}{\textbf{sky}} & \rotatebox{90}{\textbf{\textbf{person}}} & \rotatebox{90}{\textbf{rider}} & \rotatebox{90}{\textbf{car}}& \rotatebox{90}{\textbf{truck}} & \rotatebox{90}{\textbf{bus}} & \rotatebox{90}{\textbf{train}} & \rotatebox{90}{\textbf{mbike}} & \rotatebox{90}{\textbf{bike}} & \rotatebox{90}{\textbf{mIoU}} \\
        \hline
        Without Adaptation~\cite{tsai2018learning}  & 75.8 & 16.8 & 77.2 & 12.5 & 21.0 & 25.5 & 30.1 & 20.1 & 81.3 & 24.6 & 70.3 & 53.8 & 26.4 & 49.9 & 17.2 & 25.9 & \textbf{6.5} & 25.3 & 36.0 & 36.6 \\
        ROAD~\cite{chen2018road}                    & 76.3 & 36.1 & 69.6 & 28.6 & 22.4 & {28.6} & 29.3 & 14.8 & 82.3 & 35.3 & 72.9 & 54.4 & 17.8 & 78.9 & 27.7 & 30.3 & 4.0 & 24.9 & 12.6 & 39.4 \\
        AdaptSegNet~\cite{tsai2018learning}         & 86.5 & 36.0 & 79.9 & 23.4 & 23.3 & 23.9 & {35.2} & 14.8 & 83.4 & 33.3 & 75.6 & 58.5 & 27.6 & 73.7 & 32.5 & 35.4 & 3.9 & 30.1 & 28.1 & 42.4 \\
        MinEnt~\cite{vu2019advent}                  & 84.2 & 25.2 & 77.0 & 17.0 & 23.3 & 24.2 & 33.3 & \textbf{26.4} & 80.7 & 32.1 & 78.7 & 57.5 & {30.0} & 77.0 & {37.9} & 44.3 & 1.8 & 31.4 & \textbf{36.9} & 43.1 \\
        AdvEnt~\cite{vu2019advent}                  & {89.9} & {36.5} & {81.6} & {29.2} & {25.2} & {28.5} & 32.3 & 22.4 & 83.9 & 34.0 & 77.1 & 57.4 & 27.9 & {83.7} & 29.4 & 39.1 & 1.5 & 28.4 & 23.3 & 43.8 \\
        % IntraDA \cite{pan2020unsupervised}                         & {90.6} & 36.1 & {82.6} & {29.5} & 21.3 & 27.6 & 31.4 & 23.1 & {85.2} & {39.3} & {80.2} & {59.3} & 29.4 & {86.4} & 33.6 & {53.9} & 0.0 & {32.7} & {37.6} & {46.3} \\
        
        \textbf{Our BiMaL} & \textbf{91.2} & \textbf{39.6} & \textbf{82.7} & \textbf{29.4} & \textbf{25.2} & \textbf{29.6} & \textbf{34.3} & 25.5 & \textbf{85.4} & \textbf{44.0} & \textbf{80.8} & \textbf{59.7} & \textbf{30.4 }& \textbf{86.6 }& \textbf{38.5} & \textbf{47.6} & 1.2 & \textbf{34.0} & 36.8 & \textbf{47.3} \\ 
        % \textbf{Ours} & 89.1 & \textbf{38.4} & 80.9 & 26.0 & \textbf{26.1} & 27.2 & \textbf{36.9} & 17.0 & \textbf{84.5} & \textbf{37.6} & \textbf{80.1} & \textbf{59.1} & 23.1 & 82.4 & \textbf{38.2} & \textbf{52.5} & 3.9 & \textbf{36.6} & 12.2 & \textbf{44.8}  \\
        \hline 
    \end{tabular}
    }
    \label{tab:gta52city}
    \vspace{-4mm}
\end{table*}

\subsection{Network Architectures}

In our experiments, we adopt the DeepLab-V2 \cite{chen2018deeplab} with ResNet-101 \cite{He2015} backbone for the segmentation network $G$. Also, we utilize the Atrous Spatial Pyramid Pooling with sampling rate $\{6, 12, 18, 24\}$. We only use the output of layer \textit{conv5} to predict the segmentation.
In the Bijective network $F$, we use the multi-scale architecture as \cite{dinh2015nice, dinh2017density, duong2019learning, glow, Duong_2017_ICCV}. For each scale, we have multiple steps of flow, each of which consists of ActNorm, Invertible $1\times 1$ Convolution, and Affine Coupling Layer \cite{glow, 9108692}. In our experiments, the number of scales and number of flow steps are set to $4$ and $32$, respectively. 

The entire framework is implemented in PyTorch \cite{paszke2019pytorch}. Training and validating models are conducted on 4 GPUs of NVIDIA Quadpro P8000 with 48GB each GPU. Segmentation and bijective networks are trained by a Stochastic Gradient Descent optimizer \cite{bottou2010large} with learning rate $2.5\times 10^{-4}$, momentum $0.9$, and weight decay $10^{-4}$. The batch size per GPU is set to $4$ for segmentation network, and $16$ for learning bijective network. 
The image size is set to $1280 \times 720$ pixels in all experiments.
% Image sizes in all experiments are set to the resolution of $1280 \times 720$.

% \hl{On SYTHIA dataset, you should use some similar subsets such as Highway, New York-like City,  Old European Town. The ablation study should take one train and test on the other one }

\subsection{Ablation Study}

% \noindent
\textbf{Effectiveness of Losses.} Figure \ref{fig:ablation_loss} reports the semantic performance (mIoU) of BiMaL on the 16 classes of the Cityscape validation set when the model is trained on \mbox{SYNTHIA} dataset. We consider three cases: (1) without adaptation (train with source only), (2) BiMaL without regularization term ($\mathcal{L}_{llk}(\mathbf{y})$ only), and (3) BiMaL with regularization term ($\mathcal{L}_{llk}(\mathbf{y}) + \tau(\mathbf{y})$). Overall, the proposed BiMaL improve the performance of the method. In particular, the mIoU accuracy of the baseline (without adaptation) is $33.7\%$. 
% Meanwhile, 
In comparison, BiMaL without regularization and BiMaL with regularization achieve the mIoU accuracy of $43.5\%$ and $46.2\%$, respectively. 
% Analyzing 
In terms of per-class accuracy, using BiMaL significantly improves the performance on classes of \textit{`road'}, \textit{`sidewalk'}, \textit{`bus'}, and \textit{`motocycle'}.

% \noindent
\textbf{Bijective Network Ability. } 
% To be realistic and easy to visualize, 
% For a realistic and clear visualization, 
% We conduct a pilot experiment of the bijective network on an RGB image of the GTA dataset instead of segmentation images. 
% The purpose of this experiment is 
We conduct a pilot experiment of the bijective network on ground-truth semantic segmentation images of the GTA dataset.
This experiment aims to analyze the ability of the bijective network in modeling the image and structure information.
% Figure \ref{fig:rec_images} illustrates the reconstructed images of our bijective network trained on a RGB dataset. 
% In this experiment, we set up a pilot experiment to show the ability of the bijective network. 
The number of scales and number of flow steps are set to $3$, and $32$, respectively. 
As shown in Figure \ref{fig:rec_images}(a), our bijective network can successfully reconstruct good-quality images. 
% Also, our bijective network 
It also can synthesize images sampled from the latent space as shown in Figure \ref{fig:rec_images}(b). These experimental results have shown that the bijective network can model images even with complex structures as scene segmentation.

% \hl{Hau het cac bai co ve report tren 2 backbone VGG16 and Resest101 }
% \hl{Xet tren cung Resnet101, nhung bai nay hinh nhu tot hon bai cua minh}

% \hl{( Context-Aware Domain Adaptation in Semantic Segmentation - WAC 2021 }

% \hl{https://arxiv.org/pdf/2004.05498.pdf - CVPR 2020}

% \hl{DACS: Domain Adaptation via Cross-domain Mixed Sampling - WACV 2021}

\begin{figure*}[t]
    \centering
    \includegraphics[width=0.72\textwidth]{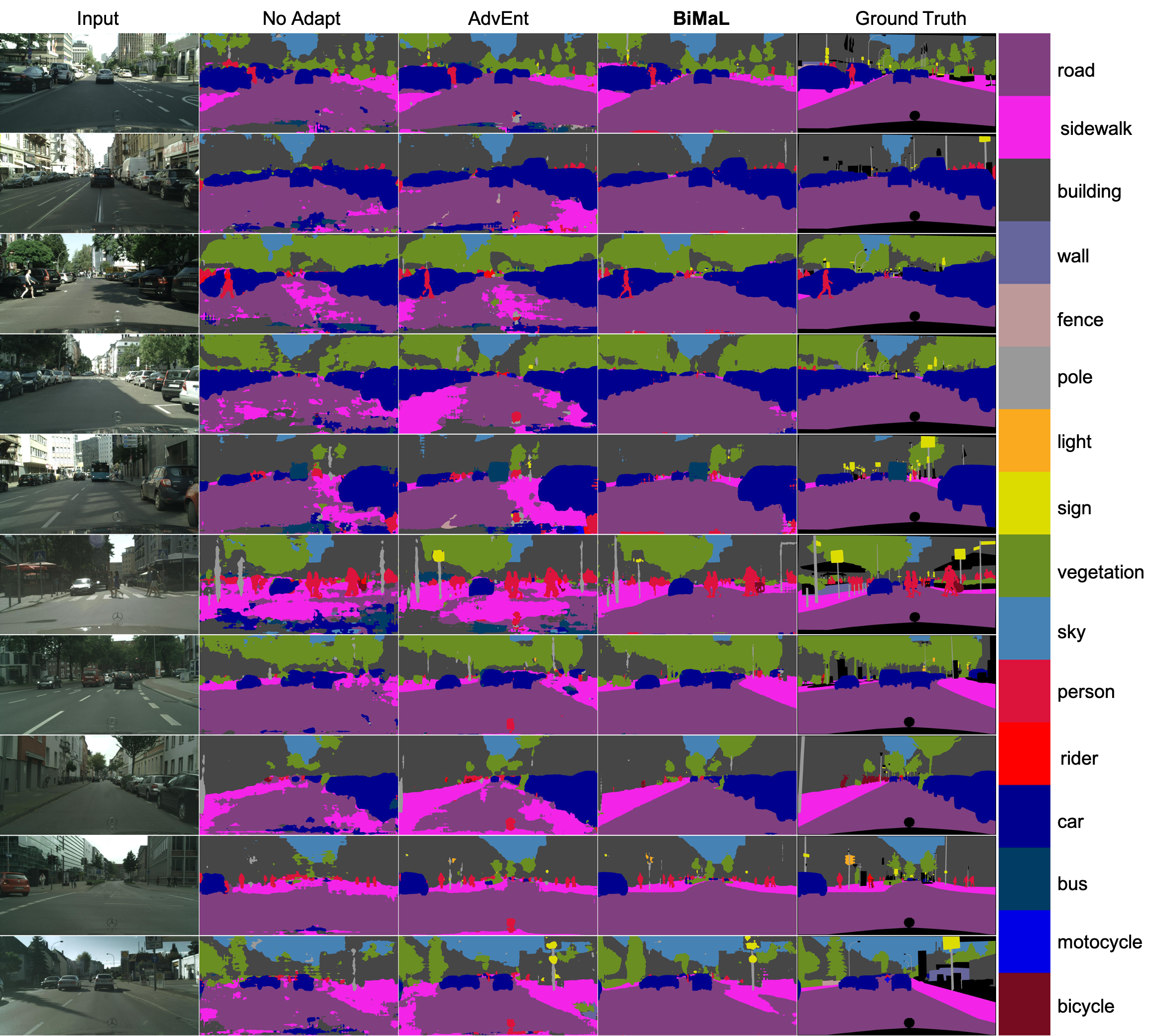}
    \caption{\textbf{Qualitative results of the SYNTHIA $\to$ Cityscapes experiment.} Columns 1 and 5 are the input and corresponding ground truth. Columns 2, 3 and 4 are the results of the model without adaptation, AdvEnt \cite{vu2019advent}, and our method.} % \textcolor{red}{[This Fig has not ref in text yet.]}}}
    \label{fig:qual_res_synthia2cityscape}
    \vspace{-5mm}
\end{figure*}

% \subsection{Results}
\subsection{Comparisons with SOTA Methods}

We present the experimental results of the proposed approach in comparison to other strong baselines.
% different baselines. 
% We consider three benchmarks in our experiments,
Comparative experiments are conducted on three benchmarks: 
i.e. SYNTHIA $to$ Cityscapes, GTA5 $to$ Cityscapes, and SYNTHIA $to$ Vistas. In all three benchmarks, our method consistently achieves the SOTA semantic segmentation performance in term of ``mean Intersection over Union" (mIoU).

%\noindent
\textbf{SYNTHIA $to$ Cityscapes.} Table \ref{tab:synthia2city} presents the semantic performance (mIoU) on the 16 classes of the Cityscape validation set. Our proposed method achieves better accuracy than the prior methods, i.e. $46.2\%$ higher than DADA \cite{vu2019dada} by $3.6\%$. Considering per-class results, our method significantly improves the results on classes of \textit{`sidewalk'} ($51.5\%$), \textit{`car'} ($85.7\%$), and \textit{`bus'} ($44.5\%$). 
We also report the results on a 13-class subset where our proposed method also achieves the State-of-the-Art performance.

% \noindent
\textbf{GTA5 $to$ Cityscapes.} Table \ref{tab:gta52city} shows the mIoU of 19 classes of Cityscapes on the validation set. Our approach gains mIoU of $47.3\%$ that is state-of-the-art performance compared to the prior methods. Analysing per-class results,
%although we cannot achieves the best results in two classes of \textit{`sign'} and \textit{`train'}, overall, 
our method gains the improvement on most classes. In particular, the results on classes of \textit{`terrain'} (+$10.0\%$), \textit{`truck'} ($+9.1\%$), \textit{`bus'} ($+8.0\%$), \textit{`motorbike'} ($+5.6\%$) demonstrate significant improvements compared to AdvEnt. 
For other classes, the proposed method gains moderate improvements, compared to prior SOTA methods.
% Other classes gain the moderate improvement compared to prior SOTA methods.
% \hl{Although our method achieves SOTA performance in almost all individual classes, there is still a limitation not only in our method but also in prior methods. Since the source dataset and the target dataset are in different countries, i.e. GTA V uses the traffic law in the US and Cityscapes uses the law in Europe, this results in the moderate performance on the class of \textit{`sign'}.}
% As shown in Table \ref{tab:gta52city}, the mIoU accuracy of 

%There are several cases, e.g. traffic lights or signs, that we may fail in some particular cases since the appearance of these objects may depend on the semantic context of the road conditions or different traffic laws. As suggestion, we will discuss in our final version of paper.

\begin{table}[b]
    \vspace{-6mm}
    \centering
    \caption{\textbf{Semantic segmentation performance mIoU (\%) on Vistas testing set of different models trained on SYNTHIA.} (const. denotes for construction)}
    % \scriptsize 
    \resizebox{8.5cm}{!}{
    \begin{tabular}{|c|c|c|c|c|c|c|c|c|}
    \multicolumn{9}{c}{SYNTHIA $\to$ Vistas (7 classes)}\\
    \hline
    Models & \rotatebox{90}{\textbf{flat}} & \rotatebox{90}{\textbf{const.}} & \rotatebox{90}{\textbf{object}} & \rotatebox{90}{\textbf{nature}} & \rotatebox{90}{\textbf{sky}} & \rotatebox{90}{\textbf{human}} & \rotatebox{90}{\textbf{vehicle}} & \rotatebox{90}{\textbf{mIoU}} \\
    \hline
     SPIGAN-no-PI~\cite{lee2018spigan}  &  53.0 & 30.8 & 3.6 & 14.6 & 53.0 & 5.8 & 26.9 & 26.8       \\
     SPIGAN~\cite{lee2018spigan}  &   74.1 & 47.1 & 6.8 & 43.3 & 83.7 & 11.2 & 42.2 & 44.1      \\
    %  AdvEnt \cite{vu2019advent}  &    82.7 & 51.8 & 18.4 & 67.8 & 79.5 & 22.7 & 54.9 & 54.0      \\
    AdvEnt \cite{vu2019advent}  & 86.9 & 58.8 & 30.5 & 74.1 & 85.1 & 48.3 & 72.5 & 65.2 \\
    DADA \cite{vu2019dada} & 86.7 & \textbf{62.1} & 34.9 & 75.9 & \textbf{88.6} & 51.1 & 73.8 & 67.6 \\
    %  \textbf{Ours}   & \textbf{87.0} &  \textbf{60.2} & \textbf{34.9} &    \textbf{76.7}    &   \textbf{86.1}  &  \textbf{52.1}     &  \textbf{73.7}       &  \textbf{67.2}  \\
    \textbf{Our BiMaL} & \textbf{87.6} & 61.6 &  \textbf{35.3} & \textbf{77.5} &  87.8 &  \textbf{53.3} & \textbf{75.6} & \textbf{68.4} \\
     \hline 
    \end{tabular}
    }
    \label{tab:synthia2vistas}
\end{table}

% \noindent
\textbf{SYNTHIA $to$ Vistas.} Table \ref{tab:synthia2vistas} reports the mIoU on 7 classes of the Vistas testing set. Our approach gains an mIoU of $67.2\%$ which is the SOTA performance compared to prior methods. 
%Except for two classes of \textit{`construction'} and \textit{`sky'}, 
Moreover, our method also gains moderate improvements in per-class accuracy.% compared to prior SOTA methods.

% \noindent
\textbf{Qualitative Results.} Figure \ref{fig:qual_res_synthia2cityscape} illustrates the qualitative results of the SYNTHIA $to$ Cityscapces experiment. %As shown in Figure \ref{fig:qual_res_synthia2cityscape}, 
Our method gives the better qualitative results compared to a model trained on the source domain and AdvEnt \cite{vu2019advent}. 
% Our results can model 
Our method can model well the structure of an image. In particular, our results have a clear border between \textit{`road'} and \textit{`sidewalk'}. Meanwhile, the results of model trained on source only and AdvEnt have an unclear border between \textit{`road'} and \textit{`sidewalk'}. Overall, our qualitative semantic segmentation results are sharper than the results of AdvEnt.

% \subsection{Ablation Study}

\vspace{-2mm}
\section{Conclusions}

This paper has presented a new Bijective Maximum Likelihood approach to domain adaptation in semantic scene segmentation. Compared to Adversarial Entropy Minimization loss, it is a more generalized form and can work without any assumption about pixel independence. 
In addition, a new Unaligned Domain Score metric has been also introduced to measure the efficiency of a segmentation model on a new target domain in the unsupervised manner. 
% Some words about experimental numbers and results ...
Through intensive experiments on three different datasets, i.e. SYNTHIA $to$ Cityscapes, GTA $to$ Cityscapes, and SYNTHIA $to$ Vistas, we achieve SOTA performance compared to prior methods. Specifically, our semantic segmentation accuracy on these three benchmarks are $46.2\%$, $47.3\%$, and $68.4\%$, respectively.
% Nevertheless, 
% There are also some challenging cases existed according to the segmentation of traffic lights or signs under different traffic laws. 
The future direction of this work is to solve challenging cases coming from the differences in ``segmentation structures'' between source and target domains such as left- and right-hand traffic.

%the segmentation of traffic lights or signs under different traffic laws or different structures (e.g. left- and right-hand traffic). 
% We leave these challenging cases as a future direction for this work.

\vspace{-4mm}
\paragraph{Acknowledgement} This work is supported by NSF Data Science, Data Analytics that are Robust and Trusted (DART), Chancellor's Innovation Fund, UAF and NSF Small Business Grant.

{\small
\bibliographystyle{ieee_fullname}
\bibliography{egpaper_final}
}

\end{document}